\documentclass{article}

\usepackage[preprint]{neurips_2026}

\usepackage[utf8]{inputenc}
\usepackage[T1]{fontenc}
\usepackage{hyperref}
\usepackage{url}
\usepackage{booktabs}
\usepackage{amsmath,amssymb,amsfonts,amsthm}
\usepackage{nicefrac}
\usepackage{microtype}
\usepackage{xcolor}
\usepackage{bm}
\usepackage{multirow}
\usepackage{enumitem}
\usepackage{algorithm}
\usepackage{algpseudocode}

\newcommand{\best}[1]{\textcolor{green!45!black}{\textbf{#1}}}
\newcommand{\second}[1]{\textcolor{blue!60!black}{\textbf{#1}}}
\newcommand{\third}[1]{\textcolor{orange!85!black}{\textbf{#1}}}

\newcommand{\R}{\mathbb{R}}
\newcommand{\E}{\mathbb{E}}
\newcommand{\Prob}{\mathbb{P}}
\newcommand{\I}{\mathbb{I}}
\newcommand{\calC}{\mathcal{C}}
\newcommand{\calA}{\mathcal{A}}
\newcommand{\calM}{\mathcal{M}}

\newcommand{\clip}{\operatorname{clip}}
\newcommand{\Normal}{\mathcal{N}}
\newcommand{\Unif}{\operatorname{Unif}}
\newcommand{\Ber}{\operatorname{Bernoulli}}

\newcommand{\CausalLongPFN}{\textsc{CausalLongPFN}}

\title{Causal Longitudinal Prior-Fitted Networks for Counterfactual Outcome Prediction}

\author{
\begin{tabular}{c}
\normalsize
Amirhossein Zare\textsuperscript{1} \quad
Amirhessam Zare\textsuperscript{1} \quad
Herlock Rahimi\textsuperscript{2}\\
\normalsize
Reza Salarikia\textsuperscript{3} \quad
Mohammad Kashkooli\textsuperscript{4,5}\\[0.45em]
\footnotesize
\texttt{amhosseinzare@gmail.com} \quad
\texttt{amir.hessam.zare@gmail.com} \quad
\texttt{herlock.rahimi@yale.edu}\\
\footnotesize
\texttt{salarikiareza@gmail.com} \quad
\texttt{mohammadkashkooli594@gmail.com}
\end{tabular}
}

\begin{document}

\maketitle
\begingroup
\renewcommand{\thefootnote}{\arabic{footnote}}
\footnotetext[1]{School of Medicine, Tehran University of Medical Sciences, Tehran, Iran.}
\footnotetext[2]{Department of Electrical and Computer Engineering, Yale University, New Haven, CT, USA.}
\footnotetext[3]{School of Medicine, Shiraz University of Medical Sciences, Shiraz, Iran.}
\footnotetext[4]{Laboratory for Computational Physiology, MIT, Cambridge, MA, USA.}
\footnotetext[5]{Student Research Committee, Shiraz University of Medical Sciences, Shiraz, Iran.}
\endgroup
\setcounter{footnote}{5}
\begin{abstract}
Longitudinal treatment decisions from multivariate time-series data require
predicting potential outcomes under future treatment sequences in the presence
of time-varying confounding, heterogeneous patient dynamics, and limited
domain-specific data. Existing longitudinal causal estimators typically address
this problem by training a new model for each cohort or simulator. We introduce
Causal Longitudinal Prior-Fitted Networks (\CausalLongPFN{}), a prior-fitted
network for time-series causal inference in longitudinal treatment-response
data and zero-shot in-context counterfactual outcome prediction. To our
knowledge, \CausalLongPFN{} is the first PFN-style model for
history-conditional potential-outcome prediction under planned longitudinal
treatment sequences, with systematic comparison against established
longitudinal causal baselines on branchable counterfactual treatment-response
benchmarks and factual real-world clinical data. The model is pretrained
entirely on synthetic episodes sampled from a broad prior over temporal
structural causal models, exposing it to treatment--confounder feedback, latent
heterogeneity, nonlinear state evolution, delayed effects, and cumulative
treatment responses. At test time, \CausalLongPFN{} remains frozen and is used
zero-shot: it conditions on support trajectories, a query history, and a
planned future treatment sequence, and returns a predictive distribution over
future outcomes without gradient updates or propensity-model fitting.
Multi-step predictions are obtained by recursively applying the one-step
predictor under the specified treatment sequence. We evaluate the model on
branchable cancer, HIV, and warfarin benchmarks with ground-truth
counterfactual labels, and on factual-only rolling-origin prediction in
MIMIC-III ICU trajectories. \CausalLongPFN{} is competitive with domain-trained
longitudinal baselines on counterfactual benchmarks and performs strongly on
factual MIMIC-III prediction, suggesting that broad synthetic causal pretraining
can provide a frozen, amortized alternative for zero-shot longitudinal
treatment-response prediction when repeated domain-specific training is costly
or impractical.
\end{abstract}

\section{Introduction}
\label{sec:introduction}

Predicting how a longitudinal outcome would evolve under future treatment
decisions is a central problem in time-series causal inference for longitudinal
treatment-response data. In the potential-outcomes framework
\citep{rubin1974estimating}, for a unit observed through
history $H_t$, the target is a history-conditional potential outcome, such as
$\mathbb{E}[Y_{t+\tau}^{(\bar a)}\mid H_t]$, under a planned treatment sequence
$\bar a=(a_t,\ldots,a_{t+\tau-1})$. Under consistency, positivity, and
sequential exchangeability, this quantity is identified by the longitudinal
$g$-formula \citep{robins1986gformula,robins2000msm,hernan2020whatif}. In
practice, however, estimating it is difficult: treatment assignment at each
step depends on covariates that are themselves affected by prior treatment (time-varying confounding); errors accumulate during multi-step counterfactual rollout; and observational cohorts are often too small
to fit reliable deep sequence models from scratch.

Modern longitudinal causal estimators address these challenges by explicitly
modeling treatment--confounder feedback and treatment effects over time. RMSN \citep{lim2018rmsn} combines
recurrent outcome models with inverse-probability weighting; CRN
\citep{bica2020crn} learns balanced recurrent representations using adversarial
treatment prediction; and G-Net \citep{li2021gnet} implements neural
$g$-computation through autoregressive simulation. More recent
transformer-based methods, including the Causal Transformer
\citep{melnychuk2022causaltransformer} and G-Transformer
\citep{xiong2024gtransformer}, use attention to represent longitudinal histories and long-range temporal dependencies and have achieved strong performance on standard counterfactual benchmarks.
Despite this progress, these methods share a fundamental operational
constraint: each new cohort or simulator typically requires a separate
supervised training run, including validation-based hyperparameter selection
and, for some methods, propensity modeling or representation balancing. This
pipeline must be repeated for every new cohort or data release.

Prior-Fitted Networks (PFNs) offer a complementary route. Rather than fitting a
new model for each dataset, a PFN is pretrained on tasks sampled from a prior
over data-generating processes and then performs in-context learning for zero-shot prediction on a new
dataset without gradient updates \citep{muller2021pfn}. This idea has led to
strong amortized predictors for tabular data \citep{hollmann2023tabpfn} and
time-series forecasting \citep{dooley2023forecastpfn,taga2025timepfn}. Recent work has also begun to apply PFN-style models for cross-sectional causal
inference, including Do-PFN, CausalPFN, and CausalFM
\citep{robertson2025dopfn,balazadeh2025causalpfn,ma2025causalfm}. However, existing causal PFNs operate on independent,
cross-sectional, tabular data: none model the sequential structure of
longitudinal histories, handle time-varying confounding, or support
multi-step potential outcomes under future treatment sequences.
CausalTimePrior \citep{thumm2026causaltimeprior} introduced a synthetic prior over temporal
SCMs with paired observational and interventional time series and demonstrated a
PFN-based proof of concept on held-out temporal SCMs. It is primarily positioned
as a generic interventional time-series prior, rather than as an end-to-end
PFN-style model for history-conditional potential-outcome prediction under
planned longitudinal treatment sequences with a systematic comparison against
established longitudinal causal baselines on synthetic and real-world
treatment-response benchmarks. The intersection of longitudinal causal
inference and PFN-style in-context prediction therefore remains largely
unexplored.

\paragraph{This work.}

We introduce \CausalLongPFN{}, a prior-fitted network for time-series causal
inference in longitudinal treatment-response data and zero-shot in-context
counterfactual outcome prediction. Given
support trajectories from a new domain, a query history observed up to time $t$,
and a supplied future treatment sequence, the frozen model returns a predictive
distribution for the query outcome under that sequence. It does so without
target-domain gradient updates, propensity-model fitting, adversarial balancing,
or domain-specific simulator access at test time.

The key idea is to amortize longitudinal causal prediction across a broad prior
over temporal structural causal models. During pretraining, each synthetic task
contains treatment-confounder feedback, latent unit heterogeneity, nonlinear
state dynamics, delayed and cumulative treatment effects, and stochastic
outcome mechanisms. The model learns to use support trajectories as an
in-context description of the task and to answer query-level potential-outcome
questions under proposed treatment sequences. At evaluation time, the learned
one-step predictor is composed autoregressively under the supplied treatment
sequence, yielding multi-step potential-outcome predictions in a zero-shot setting without retraining.

This framing does not remove the standard assumptions needed to interpret
observational data causally. Rather, \CausalLongPFN{} provides an amortized
estimator for history-conditional potential outcomes in settings where the
relevant longitudinal causal structure is supported by the synthetic prior and
the usual identification assumptions are plausible. Empirically, we compare the
frozen model against MSM, RMSN, G-Net, CRN, Causal Transformer, and
G-Transformer, each trained and tuned separately on the target domain.
\CausalLongPFN{} achieves competitive normalized RMSE on branchable
counterfactual benchmarks and strong factual prediction on MIMIC-III, suggesting
that broad synthetic causal pretraining can be a useful frozen, amortized alternative to repeated
domain-specific training.

\paragraph{Contributions.}
\begin{enumerate}[leftmargin=1.4em, itemsep=2pt, topsep=2pt]

  \item \textbf{A prior-fitted model for longitudinal causal prediction.}
  We propose \CausalLongPFN{} for history-conditional potential-outcome prediction under planned longitudinal treatment sequences. Unlike standard longitudinal causal estimators, it is evaluated as a frozen model and requires no test-time adaptation.

  \item \textbf{A synthetic prior over longitudinal causal tasks.}
  We design a temporal structural causal model prior that generates diverse
  longitudinal tasks with treatment-confounder feedback, latent unit
  heterogeneity, nonlinear lagged dynamics, delayed and cumulative treatment
  effects, regime changes, and mixed noise mechanisms. This prior supplies the support trajectories and query
  counterfactual targets used for synthetic causal pretraining.

  \item \textbf{Architecture for longitudinal in-context causal inference.}
  We propose a dual-encoder architecture combining a causal Transformer history
  encoder with a PFN context encoder over support
  trajectories and a Gaussian-mixture prediction head for distributional
  outcomes.

  \item \textbf{Autoregressive counterfactual rollout.}
  We extend the learned one-step predictor to multi-step prediction by
  autoregressively rolling it forward under supplied treatment sequences, using
  each predicted intermediate outcome as part of the subsequent query history.

  \item \textbf{Zero-shot evaluation against domain-trained baselines.}
  We evaluate a single frozen \CausalLongPFN{} on branchable cancer, HIV, and
  warfarin counterfactual benchmarks and on factual MIMIC-III ICU prediction.
  The comparison contrasts amortized synthetic pretraining with baselines that
  receive domain-specific training and validation-based model selection.

\end{enumerate}

\section{Methods}
\label{sec:methods}

This section describes \CausalLongPFN{} as an amortized model for time-series
causal inference in longitudinal treatment-response data. The method combines a
temporal structural causal prior, a support-query in-context architecture, and
autoregressive rollout under planned future treatment sequences.

\subsection{Problem formulation}
\label{sec:problem_formulation}

We consider longitudinal observational data consisting of repeated covariates,
treatments, outcomes, and static features. For unit $i$ at discrete time $t$,
let
\begin{equation}
    S_{i,t}\in\R^{d_S},\qquad
    A_{i,t}\in\calA,\qquad
    Y_{i,t}\in\R,\qquad
    C_i\in\R^{d_C}
\end{equation}
denote time-varying covariates, treatment, scalar outcome, and static
covariates. The model-facing longitudinal state is
\begin{equation}
    X_{i,t}=(S_{i,t},Y_{i,t})\in\R^{d},\qquad d=d_S+1.
\end{equation}
Implementation-specific details such as padding, the discrete four-action
interface, and inactive dimensions are described in Appendix~\ref{app:arch}.

We use the standard longitudinal ordering in which covariates and the outcome at
time $t$ are observed before treatment $A_{i,t}$ is assigned. The observed
history available at decision time $t$ is therefore
\begin{equation}
    H_{i,t}
    =
    \bigl(C_i,X_{i,0},A_{i,0},X_{i,1},\ldots,A_{i,t-1},X_{i,t}\bigr).
    \label{eq:history}
\end{equation}
A one-step potential outcome from time $t$ to $t+1$ is indexed by the candidate
treatment $A_{i,t}$ applied after observing $H_{i,t}$. Given a last observed
time $t_{\mathrm{obs}}$, a horizon $\tau\ge 1$, and
$t_\star=t_{\mathrm{obs}}+\tau$, we write the planned future treatment sequence
as
\begin{equation}
    \bar a_{t_{\mathrm{obs}}:t_\star-1}
    =
    (a_{t_{\mathrm{obs}}},a_{t_{\mathrm{obs}}+1},\ldots,a_{t_\star-1}).
\end{equation}
The first planned treatment is applied after observing $X_{t_{\mathrm{obs}}}$.

The prediction target is the conditional counterfactual predictive distribution
for a query unit,
\begin{equation}
    p\!\left(
        Y^q_{t_\star}(\bar a_{t_{\mathrm{obs}}:t_\star-1})\in dy
        \mid H^q_{t_{\mathrm{obs}}},\calC
    \right),
    \label{eq:target_predictive}
\end{equation}
where $\calC$ denotes support trajectories from the same task or domain. The support trajectories provide the in-context evidence from which the frozen
PFN infers the task-specific longitudinal data-generating process, while the
query history specifies the individual whose future potential outcome is to be
predicted. At prediction time, the model observes the
query history through $t_{\mathrm{obs}}$ and the planned future treatments.
Future query covariates are not observed under the hypothetical treatment
sequence and are therefore excluded from the query information set. For
multi-step prediction, future query outcomes are generated recursively by the
model itself.

For real observational data, a causal interpretation of
Eq.~\eqref{eq:target_predictive} requires the usual longitudinal assumptions:
consistency, positivity, and sequential exchangeability conditional on the
measured history \citep{robins1986gformula,robins2000msm,hernan2020whatif}.
Under these assumptions, the corresponding counterfactual mean is identified by
the longitudinal $g$-formula. \CausalLongPFN{} does not fit a separate
propensity model, balancing representation, or outcome model for each target
domain. Instead, it amortizes the prediction of history-conditional potential
outcomes by training a prior-fitted network on synthetic longitudinal causal
tasks sampled from a broad prior over temporal structural causal models
(TSCMs), following the structural-causal-model perspective on interventions
and counterfactuals \citep{pearl2009causality,peters2017elements}. As in prior-fitted networks, task adaptation occurs through
conditioning on the support trajectories in context, rather than through
test-time gradient updates \citep{muller2021pfn,hollmann2023tabpfn,nagler2023statistical}.

\subsection{Causal Longitudinal PFN}
\label{sec:causallongpfn}

\paragraph{Overview.}
\CausalLongPFN{} combines a synthetic prior over longitudinal causal tasks with
an in-context transformer predictor for time-series causal inference in
longitudinal treatment-response data. During pretraining, each task is sampled
from a TSCM prior and provides support trajectories together with query-level
factual or counterfactual prediction targets. After pretraining, the model is
kept frozen. At test time, it receives support trajectories from a new domain, a
query history through $t_{\mathrm{obs}}$, and a planned future treatment
sequence, and returns a predictive distribution for the query outcome under that
sequence. Thus, the model is designed as an amortized estimator for longitudinal
potential-outcome prediction rather than as a domain-specific supervised learner.

\paragraph{Temporal structural causal prior.}
\label{par:tscm_prior}

Each training episode samples a temporal structural causal model (TSCM)
$\calM\sim\Pi$ and then draws support and query trajectories from this sampled
data-generating process. The prior is designed to span a broad class of longitudinal causal dynamics rather than to reproduce a single hand-built simulator. A sampled TSCM specifies:

\begin{enumerate}[leftmargin=1.5em]
    \item \textbf{Causal temporal graph.}
    The latent longitudinal state $S_t\in\R^{d_S}$ has variable dimension and
    evolves according to sparse contemporaneous and lagged dependencies. Within a
    time slice, the instantaneous graph is acyclic; across time, lagged edges induce temporal dependence. This exposes the model to settings
    in which current covariates depend on previous covariates, previous
    treatments, and other variables in the same time slice.

    \item \textbf{Nonlinear structural mechanisms.}
    State coordinates follow sparse nonlinear autoregressive updates with
    randomly sampled elementary nonlinearities, including identity, $\tanh$,
    sinusoidal, rectified-linear, absolute-value, square, and softplus functions, and
    with Gaussian, uniform, Laplace, or zero noise. The prior therefore includes
    both smooth and nonsmooth dynamics, low- and moderate-noise regimes, and
    occasional nonstationarity through regime switches. Full sampling details are
    given in Appendix~\ref{app:prior}.

    \item \textbf{Longitudinal dynamical motifs.}
    In addition to generic nonlinear dynamics, the prior optionally overlays
    structured dynamical motifs on randomly selected state dimensions. These
    include action-memory, saturating, homeostatic, feedback-control, and
    smoothed-readout channels. The motifs are intended to capture qualitative
    mechanisms common in longitudinal data, such as delayed treatment effects,
    bounded accumulation, regulatory feedback, proxy measurements, and slow
    physiological responses. Motif equations and parameter ranges are listed in
    Appendix~\ref{app:motifs}.

    \item \textbf{Confounded behavior policy.}
    Treatments in support trajectories and factual query prefixes are sampled
    from a state-dependent stochastic behavior policy. Each unit has latent
    heterogeneity $Z_i$, which affects both its initial state and its treatment
    policy. This produces time-varying treatment--confounder feedback with
    varying strength.

    \item \textbf{Autoregressive outcome model.}
    The scalar outcome is generated as an autoregressive readout of the evolving
    state with direct and cumulative treatment effects. Consequently, the target
    may depend on current state, previous outcomes, treatment history, and
    accumulated exposure. A regime switch is included in a minority of sampled
    TSCMs to expose the model to nonstationarity.
\end{enumerate}

For interventional query episodes, the generator first simulates the query
factual prefix up to $t_{\mathrm{obs}}$. It then fixes the future treatment
sequence $\bar a_{t_{\mathrm{obs}}:t_\star-1}$ and replays the structural
equations forward from the same query state under this intervention, with future
additive noise set to its conditional mean. This produces a structural target
for the intervention-specific conditional mean. In observational episodes,
the query continues under the behavior policy and the target is factual. Details
are given in Appendix~\ref{app:oracle}.

\paragraph{Support-query episode construction.}
\label{par:episode_construction}

A pretraining episode is a supervised in-context causal prediction problem generated
from one sampled TSCM. The episode contains support trajectories, a query
trajectory prefix, a planned future treatment sequence for the query, and a
target outcome. The support trajectories serve as examples from the same
task-specific longitudinal system; the query asks for the outcome of one unit
under a factual or hypothetical continuation.

Training uses one-step prediction problems sampled at different depths along a
future path. After choosing an observation time $t_{\mathrm{obs}}$ and a future
target window, the generator samples a current rollout time
$r\ge t_{\mathrm{obs}}$ and trains the model to predict $Y^q_{r+1}$ from the
query history through $r$, the current treatment $A^q_r$, and the support
trajectories. Interventional episodes replace future behavior-policy treatments
with a sampled hypothetical sequence, whereas observational episodes retain the
factual behavior-policy continuation. Multi-step prediction is therefore not
trained as a separate direct-horizon task; it is obtained at test time by
recursively applying the learned one-step predictor.

To make the support trajectories informative about the sampled task, each
support unit contributes several labeled time points from its observed
trajectory. These labels provide in-context examples of how histories and
treatments map to subsequent outcomes within the same TSCM. Query variables
beyond the information available at the current prediction time are hidden
according to the information set defined in
Section~\ref{sec:problem_formulation}. Additional details on support-anchor
selection, masking, task-local normalization, and training augmentations are
given in Appendix~\ref{app:episode_construction}.

\paragraph{Architecture.}
\label{par:architecture}

\CausalLongPFN{} has three main components: a causal history encoder, a PFN context encoder, and a distributional prediction head.
Architectural details are provided in Appendix~\ref{app:arch}.

\emph{(i) Causal history encoder.}
A trajectory-level causal transformer, implemented using masked
self-attention in the Transformer architecture
\citep{vaswani2017attention}, maps each longitudinal sequence to
history representations. The encoder processes covariates, outcomes, treatments,
and missingness indicators while using a causal attention mask, so the
representation at time $r$ depends only on information available up to that
time. 

\emph{(ii) PFN context encoder.}
The PFN context encoder performs in-context adaptation from the support
trajectories. Support tokens summarize labeled support histories, while the
query token summarizes the query history and planned current treatment. The
support and query tokens are processed jointly by self-attention. No positional encoding is assigned based on the ordering of support trajectories, so
the architecture is designed to treat the support trajectories as an unordered
set.

\emph{(iii) Gaussian-mixture prediction head.}
The final query representation parameterizes a Gaussian mixture distribution \citep{bishop1994mixturedensity} for
the normalized next outcome,
\begin{equation}
    q_\theta(y_{r+1}\mid\calC,H^q_r,A^q_r)
    =
    \sum_{k=1}^{5}
    \pi_{r,k}\Normal(y_{r+1};\mu_{r,k},\sigma^2_{r,k}).
    \label{eq:gmm_density}
\end{equation}
The mixture head provides both a point prediction, given by the mixture mean,
and a predictive distribution for uncertainty evaluation. In the implementation,
the component means are residualized around the most recent visible or
self-predicted outcome, which gives the model a stable persistence baseline at
initialization.

\paragraph{Implementation scope.}
Our implementation of \CausalLongPFN{} uses a fixed interface across all tasks. Histories contain up to $60$ observed
time points, rollouts are evaluated up to horizon $5$, and inputs support up to
$10$ time-varying covariate channels, one scalar outcome channel, $5$ static
covariates, and four discrete treatment actions. The model uses a $4$-layer
causal history encoder and a $6$-layer PFN context encoder with hidden
dimension $256$, $8$ attention heads, feed-forward width $1024$, and a
$5$-component Gaussian-mixture prediction head, giving $8{,}138{,}384$
trainable parameters. During synthetic pretraining, each episode is generated
from a sampled TSCM and contains between $3$ and $500$ support trajectories.
With $10{,}000$ optimizer updates and effective batch size $256$, pretraining
processes $2{,}560{,}000$ independently sampled synthetic episodes. At
evaluation time, this same model is frozen and applied to all benchmark domains
without architectural changes or gradient updates. Padding, support-anchor
construction, normalization, and optimization details are provided in
Appendices~\ref{app:arch}, \ref{app:episode_construction}, and~\ref{app:training}.

\paragraph{Training and rollout.}
\label{par:training_rollout}

The model is pretrained on synthetic support-query episodes using a one-step
Gaussian-mixture negative log-likelihood. The loss is augmented with a small
auxiliary term on the mixture mean and a mild regularizer against premature
mixture collapse. Optimization details, including AdamW, learning-rate schedule,
gradient accumulation, mixed precision, gradient clipping, and stochastic PFN
depth, are reported in Appendices~\ref{app:loss} and~\ref{app:training}.

At test time, all model parameters are frozen. For one-step prediction, the
model directly evaluates
$q_\theta(y_{t_{\mathrm{obs}}+1}\mid\calC,H^q_{t_{\mathrm{obs}}},
a_{t_{\mathrm{obs}}})$. For a horizon $\tau>1$, \CausalLongPFN{} performs a
plug-in sequential rollout under the supplied treatment sequence. Starting at
$r=t_{\mathrm{obs}}$, it predicts the next-outcome distribution under the
planned treatment $a_r$, inserts the mixture mean as the next query outcome,
keeps future query covariates unavailable, and repeats this procedure until
$r=t_\star-1$. The final mixture is reported as the predictive distribution for
$Y_{t_\star}^q(\bar a_{t_{\mathrm{obs}}:t_\star-1})$, and its mean is used for
point-estimation metrics.

This rollout is a deterministic plug-in approximation to the full posterior
predictive distribution over future outcome paths. It is closely related to
sequential g-computation and parametric implementations of the longitudinal
g-formula \citep{robins1986gformula,hernan2020whatif}. The learned one-step conditional predictor is
composed forward under the specified treatment sequence, with predicted
intermediate outcomes becoming part of the subsequent query history. A stochastic
ancestral rollout that samples intermediate outcomes from the mixture is a
natural extension. Appendix~\ref{app:rollout} gives algorithmic details and discusses the
deterministic plug-in nature of this approximation.

\section{Experiments}
\label{sec:experiments}

We evaluate whether a single frozen \CausalLongPFN{} pretrained only on
synthetic TSCM episodes can serve as a zero-shot in-context predictor for time-series causal inference in external
longitudinal treatment-response data. The central comparison is between
amortized synthetic pretraining and domain-specific supervised training:
\CausalLongPFN{} is evaluated without updating its parameters, whereas all
baselines are trained and selected separately using the support trajectories of
each target domain.

\paragraph{Benchmarks.}
We use four longitudinal benchmarks: cancer tumor growth
\citep{lim2018rmsn,bica2020crn,melnychuk2022causaltransformer,li2021gnet,geng2017prediction},
warfarin PK/PD \citep{holford1986clinical,hamberg2010pkpd,iwpc2009estimation},
HIV treatment dynamics \citep{adams2004dynamic,miller2020whynot}, and
MIMIC-III ICU trajectories
\citep{PhysioNet-mimiciii-1.4,johnson2016mimic,goldberger2000physionet,wang2020mimicextract,harutyunyan2019mimicbenchmarks}.
These benchmarks are summarized in
Table~\ref{tab:evaluation_dataset_summary} and described in
Appendix~\ref{app:evaluation_datasets}. Cancer, warfarin, and HIV are
branchable simulated or semi-mechanistic systems. For these domains, the same
patient-specific dynamics can be replayed under alternative future treatment
sequences, giving ground-truth counterfactual outcomes for evaluation. MIMIC-III
is a real observational ICU dataset and is therefore used only for factual
rolling-origin prediction under the observed future treatments. Its role is to
test factual temporal prediction on real clinical trajectories, not to validate
individual counterfactual effects under unobserved interventions.

\paragraph{Evaluation configuration.}
All benchmark domains are mapped to a common longitudinal prediction format.
Each trajectory contains up to $60$ time points, prediction origins are selected
only after at least $10$ observed time points, and multi-step evaluation uses a
five-step horizon. For each domain, the evaluation grid crosses five support
sizes,
$n_{\mathrm{sup}}\in\{40,80,160,320,500\}$, ten task-index levels, and two
random repetitions, yielding $5\times10\times2=100$ benchmark tasks per domain.
Each benchmark task contains multiple rolling-origin query rows, which are first
aggregated before domain-level summaries are computed. In cancer, HIV, and
warfarin, the ten task-index levels correspond to confounding levels that
control the strength of state-dependent treatment assignment. In MIMIC-III, the
same ten-level grid is retained only to match the benchmark organization across
domains; it indexes factual rolling-origin task variants and does not alter the
observed ICU trajectories. Cancer, HIV, and warfarin provide branchable
counterfactual labels, whereas MIMIC-III provides factual labels under observed
future treatments. Dataset construction details are given in
Appendix~\ref{app:evaluation_datasets}, and scoring details are given in
Appendix~\ref{app:evaluation_protocol_details}.

\paragraph{Baselines.}
We compare against six standard longitudinal causal baselines: a marginal
structural model (MSM) \citep{robins2000msm}, Recurrent Marginal Structural
Networks (RMSN) \citep{lim2018rmsn}, G-Net \citep{li2021gnet},
Counterfactual Recurrent Networks (CRN) \citep{bica2020crn}, Causal
Transformer (CT) \citep{melnychuk2022causaltransformer}, and G-Transformer
(GT) \citep{xiong2024gtransformer}. Together, these methods represent
inverse-probability weighting, recurrent neural $g$-computation, adversarial
representation balancing, and transformer-based longitudinal counterfactual
modeling. Each baseline uses support-set validation for model selection and is
then refit on the target support trajectories before query evaluation. In
contrast, \CausalLongPFN{} receives the same support trajectories only as
in-context input and remains frozen.

\paragraph{Prediction protocol.}
All methods follow the same observation-time convention from
Section~\ref{sec:problem_formulation}. The query history is observed through
$t_{\mathrm{obs}}$, the first planned treatment is
$a_{t_{\mathrm{obs}}}$, and the target is
$Y_{t_{\mathrm{obs}}+\tau}$. For multi-step prediction, methods are evaluated
under the supplied future treatment sequence. In branchable simulated domains,
this sequence defines the intervention used to generate the counterfactual
label. In MIMIC-III, the sequence is the observed future treatment path and the
label is factual. Implementation details of scoring are given in
Appendix~\ref{app:evaluation_protocol_details}.

\paragraph{Metrics.}
The primary metric is normalized RMSE. Normalization statistics are computed
from support trajectories only, so query targets are never used to define the
reporting scale. Metrics are computed for each benchmark task by aggregating all scored query rows within that task and are then averaged within domains. Domain-balanced summaries average the
four domain means equally, preventing large domains from dominating the reported
overall performance. For MIMIC-III, normalized RMSE measures factual temporal
prediction under observed clinical practice rather than counterfactual
accuracy. Lower values are better; in result tables, the best, second-best, and
third-best values within each comparison are highlighted in green, blue, and
orange, respectively. Full scoring details are provided in
Appendix~\ref{app:evaluation_protocol_details}.

\subsection{Results}
\label{sec:results}

\paragraph{Domain-balanced performance.}
Table~\ref{tab:domain_balanced_results} reports the mean normalized RMSE after
first aggregating within each domain and then averaging equally across the four
domains. \CausalLongPFN{} achieves the best
domain-balanced one-step performance, with normalized RMSE $0.2217$, narrowly
ahead of MSM ($0.2233$) and RMSN ($0.2247$). For five-step prediction,
\CausalLongPFN{} ranks third overall, behind RMSN and G-Net, while remaining
ahead of MSM, CRN, GT, and CT. These results show that the frozen
synthetically pretrained model is competitive with baselines that are trained
and selected separately for each target domain.

\begin{table}[t]
\centering
\small
\setlength{\tabcolsep}{7pt}
\newcommand{\dm}[2]{#1{\scriptsize$\pm$#2}}
\caption{Domain-balanced normalized RMSE across cancer, HIV, MIMIC-III, and
warfarin. Entries report mean $\pm$ standard deviation across domains. Lower is
better. \CausalLongPFN{} is best for one-step prediction and third for
five-step rollout despite using no domain-specific training.}
\label{tab:domain_balanced_results}
\begin{tabular}{lcc}
\toprule
Method & One-step & Horizon-5 \\
\midrule
\CausalLongPFN{}
& \dm{\best{0.222}}{0.269}
& \dm{\third{0.389}}{0.214} \\
MSM
& \dm{\second{0.223}}{0.275}
& \dm{0.418}{0.292} \\
RMSN
& \dm{\third{0.225}}{0.273}
& \dm{\best{0.350}}{0.254} \\
G-Net
& \dm{0.247}{0.251}
& \dm{\second{0.379}}{0.223} \\
CT
& \dm{0.258}{0.259}
& \dm{0.871}{0.096} \\
GT
& \dm{0.272}{0.238}
& \dm{0.489}{0.164} \\
CRN
& \dm{0.347}{0.184}
& \dm{0.472}{0.188} \\
\bottomrule
\end{tabular}
\end{table}

\paragraph{Per-domain results.}
Table~\ref{tab:per_domain_results} reports normalized RMSE by domain, prediction
task, and method. The main pattern is that
\CausalLongPFN{} remains consistently competitive across heterogeneous domains
without target-domain retraining. For one-step prediction, it ranks second on
cancer, third on HIV, first on MIMIC-III, and second on warfarin. For five-step
prediction, it ranks first on MIMIC-III and second on warfarin, but is weaker
on HIV and cancer, where domain-trained recurrent baselines perform best. This
domain-level breakdown is important: \CausalLongPFN{} is not uniformly superior,
but it provides a strong frozen predictor across tasks with very different
dynamics and outcome scales.

\begin{table}[t]
\centering
\scriptsize
\setlength{\tabcolsep}{2.8pt}
\newcommand{\ms}[2]{#1{\scriptsize$\pm$#2}}
\caption{Per-domain normalized RMSE with standard deviation across evaluation units.
Entries report mean $\pm$ standard deviation. Lower mean normalized RMSE is
better. The top three values in each row are highlighted. MIMIC-III is
factual-only; cancer, HIV, and warfarin provide branchable counterfactual
labels.}
\label{tab:per_domain_results}
\begin{tabular}{llccccccc}
\toprule
Domain & Task & \CausalLongPFN{} & MSM & RMSN & G-Net & CRN & CT & GT \\
\midrule
Cancer & One-step
& \ms{\second{0.167}}{0.255}
& \ms{0.200}{0.278}
& \ms{\best{0.166}}{0.256}
& \ms{\third{0.168}}{0.242}
& \ms{0.251}{0.291}
& \ms{0.209}{0.265}
& \ms{0.217}{0.281} \\

Cancer & Horizon-5
& \ms{0.385}{0.356}
& \ms{0.465}{0.435}
& \ms{\best{0.246}}{0.337}
& \ms{\third{0.308}}{0.285}
& \ms{\second{0.278}}{0.334}
& \ms{0.849}{0.743}
& \ms{0.372}{0.456} \\
\midrule
HIV & One-step
& \ms{\third{0.066}}{0.032}
& \ms{\second{0.061}}{0.027}
& \ms{\best{0.051}}{0.029}
& \ms{0.097}{0.066}
& \ms{0.244}{0.175}
& \ms{0.100}{0.058}
& \ms{0.094}{0.056} \\

HIV & Horizon-5
& \ms{0.288}{0.174}
& \ms{\third{0.248}}{0.122}
& \ms{\best{0.186}}{0.117}
& \ms{\second{0.235}}{0.137}
& \ms{0.405}{0.253}
& \ms{0.915}{0.618}
& \ms{0.342}{0.193} \\
\midrule
MIMIC-III & One-step
& \ms{\best{0.617}}{0.256}
& \ms{\second{0.619}}{0.256}
& \ms{0.626}{0.272}
& \ms{\third{0.619}}{0.246}
& \ms{0.622}{0.249}
& \ms{0.638}{0.264}
& \ms{0.620}{0.251} \\

MIMIC-III & Horizon-5
& \ms{\best{0.688}}{0.198}
& \ms{0.809}{0.260}
& \ms{0.729}{0.226}
& \ms{\third{0.710}}{0.193}
& \ms{0.725}{0.214}
& \ms{0.972}{0.373}
& \ms{\second{0.694}}{0.196} \\
\midrule
Warfarin & One-step
& \ms{\second{0.036}}{0.023}
& \ms{\best{0.014}}{0.007}
& \ms{\third{0.055}}{0.085}
& \ms{0.102}{0.109}
& \ms{0.270}{0.191}
& \ms{0.084}{0.091}
& \ms{0.158}{0.188} \\

Warfarin & Horizon-5
& \ms{\second{0.196}}{0.143}
& \ms{\best{0.152}}{0.075}
& \ms{\third{0.238}}{0.236}
& \ms{0.261}{0.213}
& \ms{0.480}{0.315}
& \ms{0.749}{0.615}
& \ms{0.546}{0.646} \\
\bottomrule
\end{tabular}
\end{table}

\paragraph{Real-world factual prediction.}
MIMIC-III provides a test of transfer to real-world factual ICU trajectories,
where no method has access to counterfactual labels and evaluation is restricted
to rolling-origin prediction under observed treatment paths. \CausalLongPFN{} ranks first
on both MIMIC-III one-step and five-step prediction. For one-step prediction,
its normalized RMSE is $0.6170$, ahead of MSM at $0.6186$ and G-Net at
$0.6193$. For five-step prediction, it obtains $0.6884$, ahead of GT at
$0.6938$ and G-Net at $0.7104$. Thus, on the real clinical benchmark, the frozen
synthetically pretrained model matches or exceeds domain-trained baselines
without using target-domain gradient updates.

\paragraph{Counterfactual simulated domains.}
Cancer, HIV, and warfarin provide branchable counterfactual labels, allowing
direct evaluation under alternative treatment sequences. On one-step
counterfactual prediction, \CausalLongPFN{} is close to the strongest
domain-trained methods: it ranks second on cancer, third on HIV, and second on
warfarin. Longer-horizon performance is more mixed. \CausalLongPFN{} remains
second on warfarin and competitive on HIV, but its largest relative gap occurs
on cancer five-step prediction, where RMSN, CRN, G-Net, and GT achieve lower
error. This suggests that specialized recurrent or transformer models can retain
an advantage when a target simulator provides enough support data for
domain-specific fitting, especially over longer rollouts.

\paragraph{Interpretation.}
Overall, the results support the main claim that broad synthetic causal
pretraining can produce a useful in-context model for longitudinal
treatment-response prediction. \CausalLongPFN{} is not uniformly best, but it
achieves the best domain-balanced one-step performance, the third-best
domain-balanced five-step performance, and the best performance on the real
MIMIC-III benchmark at both horizons. These results are notable because
\CausalLongPFN{} is evaluated as a single frozen model trained only on synthetic
TSCM episodes, whereas all baselines receive domain-specific training and
validation-based model selection. The pattern suggests that a sufficiently broad
synthetic longitudinal causal prior can capture reusable structure across
treatment-response tasks, making \CausalLongPFN{} a strong general-purpose
in-context predictor when repeated domain-specific training is expensive, rapid
adaptation to a new cohort is needed, or counterfactual supervision is
unavailable.

\subsection{Uncertainty and Calibration}
\label{sec:uncertainty_calibration}

Table~\ref{tab:pfn_onestep_calibration} evaluates predictive uncertainty using
standard probabilistic forecasting diagnostics, including empirical coverage, NLL,
CRPS, and PIT-ECE
\citep{gneiting2007probabilistic,gneiting2007strictly}. Calibration varies by domain. Warfarin has the lowest RMSE, NLL, and CRPS, and
its empirical coverage is slightly conservative at the $90\%$ level. HIV also
shows low point error and sharp predictive intervals, although coverage remains
below nominal. MIMIC-III is the most difficult calibration setting: it has the
largest NLL and CRPS and substantially wider intervals, reflecting the greater
heterogeneity and noise of the real ICU benchmark. Overall, the Gaussian-mixture
head provides useful distributional information without domain-specific
training, but the under-coverage suggests that future work should improve
uncertainty propagation, especially for multi-step prediction and real-world
clinical data.

\begin{table}[t]
\centering
\scriptsize
\setlength{\tabcolsep}{3pt}
\caption{One-step calibration of \CausalLongPFN{} predictive distributions.
Lower is better for RMSE, NLL, CRPS, and PIT-ECE. Empirical coverage should
match the nominal level, while interval width should be interpreted relative to
coverage.}
\label{tab:pfn_onestep_calibration}
\begin{tabular}{lrrrrrrrrr}
\toprule
Domain & RMSE & NLL $\downarrow$ & CRPS $\downarrow$ & Pred. std. & PIT-ECE $\downarrow$ & Cov. 80\% & Width 80\% & Cov. 90\% & Width 90\% \\
\midrule
Cancer    & 0.167 & -0.711 & 0.082 & 0.054 & 0.029 & 0.703 & 0.125 & 0.781 & 0.170 \\
HIV       & 0.066 & -1.310 & 0.039 & 0.064 & 0.037 & 0.753 & 0.155 & 0.849 & 0.205 \\
MIMIC-III & 0.617 &  0.938 & 0.336 & 0.451 & 0.019 & 0.727 & 1.096 & 0.836 & 1.518 \\
Warfarin  & 0.036 & -1.976 & 0.021 & 0.048 & 0.036 & 0.863 & 0.106 & 0.934 & 0.145 \\
\midrule
Domain-balanced & 0.222 & -0.765 & 0.120 & 0.154 & 0.030 & 0.761 & 0.370 & 0.850 & 0.510 \\
\bottomrule
\end{tabular}
\end{table}

\section{Conclusion}
\label{sec:conclusion}

We introduced \CausalLongPFN{}, a prior-fitted transformer for predicting history-conditional potential outcomes in longitudinal treatment-response
time series. The model is pretrained solely on synthetic temporal structural causal
models and is then evaluated as a frozen in-context predictor on new domains.
Given support trajectories, a query history, and a planned future treatment
sequence, it returns a predictive distribution without target-domain gradient
updates, propensity-model fitting, adversarial balancing, or simulator access at
test time.

\CausalLongPFN{} achieves the best domain-balanced one-step normalized RMSE and
the third-best domain-balanced five-step normalized RMSE. It performs
particularly well on factual MIMIC-III rolling-origin prediction, where it ranks
first at both horizons.

These results suggest that broad synthetic causal pretraining can provide a useful in-context predictor for time-series causal inference and longitudinal
treatment-response tasks, especially when retraining is costly, rapid evaluation
on a new cohort is needed, or counterfactual supervision is unavailable. At the
same time, the results show that domain-specific training remains valuable when
sufficient target-domain data and validation signal are available.

\section{Limitations and broader impact}
\label{sec:limitations_broader_impact}

As a model for time-series causal inference in longitudinal treatment-response
data, \CausalLongPFN{} does not remove the assumptions required for causal
interpretation of longitudinal observational data. In real cohorts,
counterfactual validity still depends on consistency, positivity, and
sequential exchangeability given the measured history, as summarized in
Appendix~\ref{app:identification-assumptions}. Violations due to unmeasured
confounding, poor treatment overlap, censoring,
irregular sampling, or measurement error can bias any longitudinal
counterfactual estimator, including \CausalLongPFN{}.

The method also depends on the support of the synthetic prior. Performance may
degrade when the target domain contains mechanisms, treatment policies, outcome
dynamics, missingness patterns, or intervention effects that are poorly covered
by the TSCM prior. The current implementation focuses on discrete treatments,
fixed time grids, and deterministic mean rollout, which make the model stable,
efficient, and straightforward to evaluate across heterogeneous benchmarks.
These choices are not fundamental restrictions of the framework. Natural
extensions include continuous or structured treatment spaces, irregular-time
encoders, explicit missingness and censoring models, and stochastic rollout
procedures that propagate uncertainty over future trajectories while preserving
the same amortized in-context causal prediction principle.

The potential benefit of this approach is to reduce dependence on hand-built
disease simulators and repeated domain-specific supervised training when
studying longitudinal treatment-response prediction from time-series data. A frozen in-context model
could be useful for rapid benchmarking, exploratory counterfactual analysis, or
settings where retraining many specialized models is impractical. The main risk
is over-reliance: predictions may appear precise even when the causal assumptions,
data quality, treatment overlap, or prior support are inadequate. In particular,
strong factual prediction on MIMIC-III should not be interpreted as validation
of individual treatment effects under unobserved ICU interventions.
\CausalLongPFN{} should therefore be viewed as a research tool for causal
sequence modeling and hypothesis generation, not as a standalone clinical
decision system.

\section*{Code availability}

Code for model training, synthetic episode generation, benchmark construction,
and evaluation is available at
\url{https://github.com/Amirhossein-Zare/causal-long-pfn}.

\section*{Model availability}

The pretrained \CausalLongPFN{} inference weights are available on the Hugging
Face Model Hub at
\url{https://huggingface.co/Amirhossein-Zare/causal-long-pfn}.
The released checkpoint,
\texttt{causal-long-pfn-v1-step10000.safetensors}, contains the frozen model
weights used for inference.

\section*{Data availability}

The cancer, HIV, and warfarin benchmarks are simulated or semi-mechanistic
benchmarks that can be regenerated using the released code and the simulator
specifications described in the paper. MIMIC-III is a credentialed-access
de-identified clinical database and is not redistributed with this paper.
Reproducing MIMIC-III experiments requires obtaining access through the official
data-use process and applying the preprocessing protocol described in
Appendix~\ref{app:data_mimic}.

\section*{Funding}

No external funding was received for this work.

\section*{Competing interests}

The authors declare no competing interests.

\bibliographystyle{plainnat}
\bibliography{references}

\newpage
\appendix

\section{Causal foundations and estimand}
\label{app:causal-foundations}

This appendix states the longitudinal causal estimand used in the paper and the
standard assumptions under which it can be interpreted causally from
observational data.

\subsection{Observed data and histories}
\label{app:obs-data-histories}

For unit $i$ at discrete time $t$, let
\begin{equation}
    S_{i,t}\in\R^{d_S},\qquad
    A_{i,t}\in\calA,\qquad
    Y_{i,t}\in\R,\qquad
    C_i\in\R^{d_C}
\end{equation}
denote time-varying covariates, treatment, scalar outcome, and static
covariates. The model-facing longitudinal state is
\begin{equation}
    X_{i,t}=(S_{i,t},Y_{i,t})\in\R^d,\qquad d=d_S+1.
\end{equation}
Treatment $A_{i,t}$ is assigned after observing $X_{i,t}$. The observed history
available immediately before treatment assignment at time $t$ is
\begin{equation}
    H_{i,t}
    =
    (C_i,X_{i,0},A_{i,0},X_{i,1},A_{i,1},\ldots,A_{i,t-1},X_{i,t}).
    \label{eq:history-def}
\end{equation}

For a query unit observed through time $t$, the model receives support
trajectories $\calC$ from the same task or domain, the query history $H_t$, and
a planned future treatment sequence
\begin{equation}
    \bar a_{t:t+\tau-1}
    =
    (a_t,a_{t+1},\ldots,a_{t+\tau-1}).
\end{equation}
The target is the history-conditional potential outcome
\begin{equation}
    Y_{t+\tau}(\bar a_{t:t+\tau-1}),
\end{equation}
or its conditional predictive distribution given the observed query history and
support trajectories:
\begin{equation}
    p\!\left(
        Y_{t+\tau}(\bar a_{t:t+\tau-1})\in dy
        \mid H_t,\calC
    \right).
    \label{eq:appendix-target}
\end{equation}
For point prediction, we evaluate the corresponding conditional mean.

\subsection{Identification assumptions}
\label{app:identification-assumptions}

For observational data, Eq.~\eqref{eq:appendix-target} has a causal
interpretation only under the usual longitudinal causal assumptions.

\paragraph{Consistency.}
If a unit actually follows the treatment sequence
$\bar a_{t:t+\tau-1}$, then its observed outcome equals the corresponding
potential outcome under that sequence.

\paragraph{Sequential exchangeability.}
At each time point, after conditioning on the observed history $H_t$, treatment
assignment is independent of future potential outcomes. Informally, there are no
unmeasured time-varying confounders after conditioning on the recorded history.

\paragraph{Positivity.}
Every treatment sequence considered for evaluation must have positive
probability, or adequate support, among units with comparable histories. Without
such overlap, the corresponding counterfactual prediction requires extrapolation.

\paragraph{No interference and well-defined interventions.}
One unit's potential outcomes are unaffected by the treatment assignments of
other units, and the treatment actions correspond to well-defined interventions.

These assumptions are standard for longitudinal causal inference and are not
guaranteed by \CausalLongPFN{}. They are required for any observational
counterfactual interpretation of the predictions.

\subsection{Connection to the longitudinal $g$-formula}
\label{app:gformula-identification}

Under consistency, sequential exchangeability, positivity, and no interference,
the conditional mean potential outcome can be written using the longitudinal
$g$-formula \citep{robins1986gformula,robins2000msm,hernan2020whatif}. In
words, the $g$-formula propagates the observed conditional transition law
forward while setting future treatments to the specified intervention sequence.

Let
\begin{equation}
    K_s(dx_{s+1}\mid h_s,a_s)
    =
    \Prob(X_{s+1}\in dx_{s+1}\mid H_s=h_s,A_s=a_s)
\end{equation}
denote the observed one-step transition distribution. Starting from
$h_t=H_t$, define the future history recursively by appending the intervention
treatment $a_s$ and the next state $x_{s+1}$. Then the identified conditional
mean can be written schematically as
\begin{equation}
    \E\!\left[
        Y_{t+\tau}(\bar a_{t:t+\tau-1})
        \mid H_t=h_t
    \right]
    =
    \int
    y(x_{t+\tau})
    \prod_{s=t}^{t+\tau-1}
    K_s(dx_{s+1}\mid h_s,a_s).
    \label{eq:appendix-gformula}
\end{equation}
This expression motivates the sequential prediction problem studied in the
paper: future outcomes are predicted by repeatedly applying one-step conditional
models under a specified future treatment sequence.

\subsection{Role of \CausalLongPFN{}}
\label{app:pfn-role}

\CausalLongPFN{} is an estimator for the prediction problem above. It does not
introduce new identification assumptions and does not remove the need for
consistency, positivity, and sequential exchangeability in observational data.
Instead, it amortizes the estimation problem by pretraining on many synthetic
longitudinal causal tasks and then conditioning on support trajectories from a
new task at test time.

The model is trained as a one-step predictor. Multi-step predictions are
obtained by deterministic plug-in rollout: the model predicts the next outcome
under the planned treatment, inserts the predicted mean into the query history,
and repeats this process until the desired horizon. This procedure is an
approximation to full sequential predictive inference because it does not
integrate over all possible intermediate outcome paths. The empirical evaluation directly assesses the resulting multi-step predictions in
the benchmark settings.

\section{Temporal structural causal prior}
\label{app:prior}

This appendix specifies the temporal structural causal model (TSCM) prior used
to generate synthetic pretraining episodes for \CausalLongPFN{}. Each episode
draws a fresh longitudinal data-generating process from the prior and then
samples support trajectories and a query trajectory from that process. The prior
is intentionally heterogeneous: it varies state dimension, temporal lag
structure, nonlinear mechanisms, treatment-policy confounding, latent unit
heterogeneity, outcome dynamics, observation windows, and interventional
rollout horizons. Its purpose is not to reproduce any single disease simulator,
but to expose the model to a broad family of longitudinal treatment-response
tasks with reusable causal structure.

\subsection{Global ranges}

\begin{table}[h]
\centering
\caption{Core synthetic task ranges. The prior varies state dimension, support
size, observation time, and prediction horizon so that a single model is trained
across heterogeneous longitudinal causal tasks.}
\label{tab:data_hparams}
\begin{tabular}{ll}
\toprule
Quantity & Value \\
\midrule
Observed time $t_{\mathrm{obs}}$ & Uniform integer $1$--$60$ \\
Prediction horizon $\tau$ & Uniform integer $1$--$5$ \\
Maximum input length & $65$ input slots, target index up to $65$ \\
State dimension $d_S$ & Uniform integer $1$--$10$ \\
Outcome dimension & $1$ \\
Padded input dimension $D_{\max}$ & $11$ \\
Static covariate dimension & $5$; active in $30\%$ of synthetic episodes \\
Treatment space & $4$ discrete treatments \\
Latent heterogeneity dimension & $3$ \\
Support size & Uniform integer $3$--$500$ \\
Support anchor labels & $4$ per support trajectory \\
Observational query probability & $0.30$ \\
Support future-covariate masking probability & $0.35$ \\
Support target-noise augmentation & $0.15$ \\
Sentinel for hidden values & $-99$ \\
\bottomrule
\end{tabular}
\end{table}

\subsection{TSCM hyperparameter sampling}
\label{app:dgp_hypers}

A synthetic TSCM instance is sampled as follows:
\begin{enumerate}[leftmargin=1.5em]
    \item \textbf{State dimension.} Sample
    $d_S\sim\Unif\{1,\ldots,10\}$.

    \item \textbf{Lag order.} Sample $K\sim\Unif\{1,2\}$.

    \item \textbf{Instantaneous graph.} Sample an instantaneous adjacency matrix
    $G^{(0)}\in\{0,1\}^{d_S\times d_S}$ as a strictly lower-triangular
    Erdős--Rényi matrix with edge probability
    \begin{equation}
        p_{\mathrm{edge}}=0.1+0.5B,\qquad B\sim\operatorname{Beta}(2,2).
    \end{equation}
    This gives an acyclic contemporaneous graph under the coordinate ordering.

    \item \textbf{Lagged graph.} For lag $k$, sample a full lagged adjacency
    matrix $G^{(k)}\in\{0,1\}^{d_S\times d_S}$ with edge probability
    $p_{\mathrm{edge}}\gamma_{\mathrm{lag}}^k$, where
    $\gamma_{\mathrm{lag}}\sim\Unif(0.4,0.8)$. This induces sparse temporal
    dependence with geometrically decaying edge probability across lags.

    \item \textbf{Structural weights.} Sample instantaneous weights
    $W^{(0)}_{ij}\sim\Normal(0,\sigma_W^2)G^{(0)}_{ij}$ and lagged weights
    $W^{(k)}_{ij}\sim\Normal(0,(0.7\sigma_W)^2)G^{(k)}_{ij}$, with
    $\sigma_W\sim\Unif(0.3,1.0)$.

    \item \textbf{Nonlinearities.} Sample each generic activation independently
    from
    \begin{equation}
        \{\mathrm{id},\tanh,\sin,\cos,|\cdot|,(\cdot)^2,
        \operatorname{ReLU},\operatorname{softplus}\}.
    \end{equation}

    \item \textbf{State noise.} For each state coordinate, sample a centered
    Gaussian, uniform, or Laplace noise family. The coordinate noise scale is
    zero with probability $0.5$; otherwise it is proportional to a task-level
    base scale. The base scale is sampled from a low-noise range with
    probability $0.6$ and from a moderate-noise range with probability $0.4$.

    \item \textbf{Autoregressive persistence.} Set the coordinate-level
    autoregressive coefficient $\alpha_i=0$ with probability $0.5$; otherwise
    sample $\alpha_i\sim\Unif(0.5,1.0)$.

    \item \textbf{Treatment-policy confounding strength.} Sample a policy
    strength multiplier that is zero with probability $0.08$, one with
    probability $0.20$, and otherwise a random integer from $2$ to $5$.
    Treatment-policy state weights are scaled by this multiplier, producing
    tasks with varying degrees of treatment--confounder feedback.

    \item \textbf{Regime switch.} With probability $0.12$, sample a second
    structural mechanism with the same graph support and activate it after a
    sampled early-to-midpoint switch time. Structural weights and nonlinearities
    change after the switch, creating nonstationary longitudinal dynamics.
\end{enumerate}

\subsection{Generic structural mechanisms}

For a generic non-motif state coordinate, the transition is a sparse nonlinear
autoregressive update combining lagged state inputs, acyclic within-slice
inputs, treatment inputs, and additive noise. Instantaneous contributions use
the partially constructed next-time state $S_{t+1,\ell}$ for $\ell<m$, while
lagged contributions use previous states from the lag buffer. Values are clipped
internally to avoid numerical explosions during synthetic generation. Thus, even
before adding the structured motifs below, the prior spans nonlinear
autoregression, contemporaneous acyclic dependence, lagged temporal dependence,
treatment effects, and heterogeneous noise.

\subsection{Latent dynamical motifs}
\label{app:motifs}

The prior optionally allocates disjoint state coordinates to five motif types.
Motif coordinates are selected by a random permutation of the state dimensions,
so motif identity is not tied to a fixed input channel. These motifs are
included to expose the model to qualitative mechanisms common in biomedical and
behavioral longitudinal data: slow accumulation, saturation, homeostatic
regulation, feedback control, and proxy readout dynamics.

\paragraph{Action-memory channel.}
With probability $0.25$, one coordinate follows a leaky accumulation model:
\begin{equation}
    S_{t+1,m}^{\mathrm{mem}}
    =
    \delta_m S_{t,m}^{\mathrm{mem}}
    +w_m^\top b(A_t)
    +v_m^\top M_{t+1}
    +\varepsilon_{t+1,m},
\end{equation}
where $\delta_m\sim\Unif(0.72,0.97)$ and $M_{t+1}$ is a running treatment-memory
vector.

\paragraph{Saturating channel.}
With probability $0.25$, one or two coordinates follow a nonnegative saturating
update:
\begin{equation}
    S_{t+1,m}^{\mathrm{sat}}
    =
    \clip_{[0,6]}\!\left(
        S_{t,m}^{\mathrm{sat}}
        +r_mb_m\left(1-g_m\frac{L_t}{h_m+L_t+\epsilon}\right)
        -r_mS_{t,m}^{\mathrm{sat}}
        +\varepsilon_{t+1,m}
    \right),
\end{equation}
where $L_t$ is a nonnegative signal derived from treatment memory and, when
available, latent memory coordinates.

\paragraph{Homeostatic channel.}
With probability $0.25$, one coordinate reverts toward a sampled baseline:
\begin{equation}
    S_{t+1,m}^{\mathrm{hom}}
    =
    S_{t,m}^{\mathrm{hom}}
    +\kappa_m(\mu_m-S_{t,m}^{\mathrm{hom}})
    +w_m^\top b(A_t)
    +\varepsilon_{t+1,m}.
\end{equation}

\paragraph{Feedback channel.}
With probability $0.25$, one coordinate receives error-driven control from a
source coordinate $j(m)$:
\begin{equation}
    S_{t+1,m}^{\mathrm{fb}}
    =
    \rho_m S_{t,m}^{\mathrm{fb}}
    +\gamma_m(\eta_m-S_{t,j(m)})
    +w_m^\top b(A_t)
    +\varepsilon_{t+1,m}.
\end{equation}

\paragraph{Readout channel.}
With probability $0.20$, one coordinate tracks another coordinate using
exponential smoothing:
\begin{equation}
    S_{t+1,m}^{\mathrm{read}}
    =
    \rho_m^{\mathrm{rd}}S_{t,m}^{\mathrm{read}}
    +(1-\rho_m^{\mathrm{rd}})S_{t+1,j(m)}
    +\varepsilon_{t+1,m}.
\end{equation}

\begin{table}[h]
\centering
\caption{Sampling ranges for motif-specific parameters. The motifs introduce
slow accumulation, saturation, regulation, feedback, and proxy readout dynamics
into the synthetic prior.}
\label{tab:motif_params}
\small
\begin{tabular}{llll}
\toprule
Motif & Parameter & Symbol & Range \\
\midrule
Memory & decay & $\delta_m$ & $[0.72,0.97]$ \\
Saturating & baseline & $b_m$ & $[0.5,1.5]$ \\
 & rate & $r_m$ & $[0.02,0.15]$ \\
 & gain & $g_m$ & $[0.25,0.95]$ \\
 & half-saturation & $h_m$ & $[0.3,2.0]$ \\
Homeostatic & reversion & $\kappa_m$ & $[0.03,0.15]$ \\
 & baseline & $\mu_m$ & $[-0.5,0.5]$ \\
Feedback & decay & $\rho_m$ & $[0.65,0.95]$ \\
 & gain & $\gamma_m$ & $[0.10,0.90]$ \\
Readout & smoothing & $\rho_m^{\mathrm{rd}}$ & $[0.70,0.97]$ \\
\bottomrule
\end{tabular}
\end{table}

\subsection{Latent heterogeneity and behavior policy}
\label{app:latent}

Unit heterogeneity is encoded by a latent vector
$Z_i\sim\Normal(0,I_3)$ drawn once per support or query trajectory. This latent
factor affects both the initial state and the treatment policy:
\begin{align}
    S_{i,0} &\approx U_{S0}Z_i + \varepsilon_{i,0},\\
    W_{u,i} &= W_u + U_u Z_i,\qquad u\in\{0,1\}.
\end{align}
Treatment memories used by the policy evolve as
\begin{equation}
    M_{t+1,u}
    =
    \lambda_u M_{t,u}+b_u(A_t),
    \qquad
    \lambda_u\sim\Unif(0.5,0.95).
\end{equation}
The behavior-policy logits depend on current state, recent treatment memory,
static covariates when active, and latent heterogeneity. Because both baseline
state and treatment assignment depend on $Z_i$, and because treatment assignment
also depends on the evolving state, support trajectories exhibit persistent
unit-level heterogeneity and time-varying confounding. A small probability of
near-random policy strength preserves overlap.

\paragraph{Synthetic treatment encoding.}
In the synthetic TSCM prior, the four-valued treatment is generated through two
binary policy components. Conditional on the current state, recent treatment
memories, static covariates when active, and latent heterogeneity, the generator
computes two logistic probabilities and samples
\begin{equation}
    A_{t,0}\sim\Ber(p_{t,0}),\qquad
    A_{t,1}\sim\Ber(p_{t,1}).
\end{equation}
The treatment supplied to the model is then
\begin{equation}
    A_t=A_{t,0}+2A_{t,1}\in\{0,1,2,3\}.
\end{equation}
This bitwise construction is used only to generate heterogeneous treatment
policies. The model itself receives the resulting four-valued treatment.

Observed static covariates are included in a random subset of synthetic
episodes. Specifically, the implementation activates the five-dimensional
static covariate vector with probability $0.30$ and otherwise supplies zeros.
This prevents the model from assuming that static covariates are informative in
every task while still exposing it to domains where baseline features are
useful.

\subsection{Outcome mechanism}

The base readout $R_t$ is either a selected state coordinate or an affine
projection of all state variables. Coordinates belonging to dynamical motifs are
sampled with elevated probability as direct outcome coordinates. The scalar
outcome evolves according to an autoregressive readout with
$\rho_Y\sim\Unif(0.35,0.90)$, state gain in $[0.35,1.20]$, small direct and
cumulative treatment effects, and weak linear trends. Outcome noise is low in
most TSCMs but can be moderate in a minority of cases. Consequently, the target
may depend on current state, prior outcomes, recent treatment, and accumulated
treatment exposure.

\subsection{Counterfactual oracle construction}
\label{app:oracle}

For each interventional training example, the counterfactual target is
constructed by structural replay:
\begin{enumerate}[leftmargin=1.5em]
    \item The query trajectory is simulated under the observational behavior
    policy from $t=0$ to $t_{\mathrm{obs}}$, storing the state, treatment
    memories, and lag buffer at $t_{\mathrm{obs}}$.
    \item From $t_{\mathrm{obs}}$, a second rollout is performed under the
    hypothetical treatment sequence
    $\bar a_{t_{\mathrm{obs}}:t_\star-1}$, with future additive state noise set
    to its mean.
    \item The oracle outcome is continued from the observed outcome prefix
    $Y_{0:t_{\mathrm{obs}}}$ using the counterfactual state path and the planned
    treatments, again with future outcome noise set to its mean.
\end{enumerate}
This construction produces a conditional structural target given the factual
query history and the planned intervention. It focuses training on the mean
causal response to the supplied treatment sequence rather than on aleatoric
future noise, while stochasticity remains present in support trajectories and
factual query prefixes.

\subsection{Support anchor time points}
\label{app:anchors}

Each synthetic support trajectory contributes $K_{\mathrm{sup}}=4$ labeled
outcome anchors. During one-step pretraining, the first anchor is the current
label time $r+1$. The remaining anchors are the earliest post-observation label
time $t_{\mathrm{obs}}+1$, a midpoint between $t_{\mathrm{obs}}+1$ and $r+1$,
and a random anchor sampled from
$\{t_{\mathrm{obs}}+1,\ldots,r+1\}$. This multi-anchor strategy provides
in-context examples at several rollout depths from the same sampled TSCM, rather
than relying on a single labeled support time point.

For external benchmark evaluation, support anchors are chosen from each support
row's available outcome times using the same four-anchor interface but a
deterministic template: latest valid outcome time, midpoint, earliest valid
outcome time, and one random valid anchor. Thus, the architectural interface is
shared across pretraining and evaluation---multiple labeled support anchors per
support trajectory---while the exact anchor-selection rule is adapted to the
available benchmark rows.

\subsection{Data augmentation}
\label{app:aug}

Three augmentations are applied during synthetic training:
\begin{enumerate}[leftmargin=1.5em]
    \item \textbf{Observational mode} with probability $0.30$: the query target
    is factual under the behavior policy rather than interventional. This keeps
    ordinary factual prediction within the training distribution.

    \item \textbf{Support target noise} with probability $0.15$: the first
    support target anchor may receive additive noise at scale $0\%$, $5\%$, or
    $10\%$ of the task outcome standard deviation. This improves robustness to
    noisy support labels while leaving the other support anchors unchanged.

    \item \textbf{Support future-covariate masking} with probability $0.35$:
    support state values after $t_{\mathrm{obs}}$ and before the target horizon
    are replaced with the sentinel value while support outcome labels remain
    visible. This discourages reliance on post-intervention covariates that are
    unavailable for the query under hypothetical treatment sequences.
\end{enumerate}

\section{Training episode construction}
\label{app:episode_construction}

\begin{algorithm}[h]
\caption{Synthetic \CausalLongPFN{} episode}
\label{alg:episode}
\begin{algorithmic}[1]
\State Sample a TSCM $\calM\sim\Pi$ and support size $n_{\mathrm{ctx}}$.
\State Sample $t_{\mathrm{obs}}$, $\tau$, and
$t_\star=t_{\mathrm{obs}}+\tau$.
\State Generate $n_{\mathrm{ctx}}$ support trajectories under the behavior
policy through $t_\star$.
\State Generate one query factual prefix under the behavior policy through
$t_{\mathrm{obs}}$.
\If{interventional mode}
    \State Sample hypothetical future treatments
    $\bar a_{t_{\mathrm{obs}}:t_\star-1}$.
    \State Clone the query state, treatment memories, and lag buffer at
    $t_{\mathrm{obs}}$.
    \State Simulate future state and outcome under the intervened structural
    equations with future noise set to its mean.
\Else
    \State Continue the query under the behavior policy and use its factual
    future.
\EndIf
\State Compute task-local support normalizers; normalize, clip, pad, and mask
unavailable values.
\State Sample current time
$r\sim\Unif\{t_{\mathrm{obs}},\ldots,t_\star-1\}$ and set label
$z=Y^q_{r+1}$.
\State Choose four support anchor times: $r+1$, $t_{\mathrm{obs}}+1$, a
midpoint, and a random anchor from
$\{t_{\mathrm{obs}}+1,\ldots,r+1\}$.
\State Return support sequences, support labels, anchor times, the query sequence, treatments, current time $r$, and normalized label $z$.
\end{algorithmic}
\end{algorithm}

\paragraph{Observed-prefix training and recursive evaluation.}
During training, the model conditions on the query outcome history through the
sampled current time $r$ and predicts the next outcome $Y^q_{r+1}$; all later
query outcomes remain hidden. At evaluation time, only the query history through
$t_{\mathrm{obs}}$ is observed. For horizons beyond one step, the model performs
plug-in sequential rollout, inserting each predicted mixture mean into the query
outcome channel before predicting the next time point.

\paragraph{Normalization.}
State normalizers use only support times $0{:}t_{\mathrm{obs}}$:
\begin{equation}
    \mu_{S,m}
    =
    \operatorname{mean}_{j,t\le t_{\mathrm{obs}}}S_{j,t,m},
    \qquad
    \sigma_{S,m}
    =
    \max\{\operatorname{sd}_{j,t\le t_{\mathrm{obs}}}(S_{j,t,m}),0.1\}.
\end{equation}
Outcome normalizers use support outcomes over $1{:}t_\star$:
\begin{equation}
    \mu_Y
    =
    \operatorname{mean}_{j,1\le t\le t_\star}Y_{j,t},
    \qquad
    \sigma_Y
    =
    \max\{\operatorname{sd}_{j,1\le t\le t_\star}(Y_{j,t}),0.1\}.
\end{equation}
Episodes with near-constant outcome scale are rejected. Normalized state values
are clipped to $[-3,3]$, normalized outcomes to $[-10,10]$, and unavailable
values are marked with the sentinel $-99$.

\section{Model architecture details}
\label{app:arch}

\subsection{State encoder}
\label{app:state_enc}

Let $V_t\in\R^{D_{\max}}$ denote the padded model input at time $t$, containing
the time-varying covariates and scalar outcome, and let
$m_t=\I\{V_t<-90\}$ denote the hidden-value mask induced by the sentinel. Hidden
entries are set to zero before projection, while the mask itself is retained as
an input feature. First differences are scaled by $0.5$ and set to zero whenever
either adjacent value is hidden.

Let $y$ denote the active outcome coordinate. The encoder separates covariate
and outcome channels:
\begin{align}
    e_t^{S} &= W_S[V_t^{(-y)},0.5\Delta V_t^{(-y)},-2m_t^{(-y)}],\\
    e_t^{Y} &= W_Y[V_t^{(y)},0.5\Delta V_t^{(y)},-2m_t^{(y)}],\\
    e_t^{A} &= W_A\operatorname{onehot}(A_t).
\end{align}
The encoded time-step representation is
\begin{equation}
    e_t=\operatorname{LN}(e_t^S+e_t^Y+e_t^A).
\end{equation}
Separating the outcome channel from the covariate channels helps preserve the
distinction between observed predictors and the target process. Padded inactive
dimensions remain zero after cleaning, and the input scale factor
$\sqrt{D_{\max}/d}$ helps keep signal magnitudes comparable across tasks with
different active state dimensions.

\subsection{History encoder}
\label{app:hist_enc}

The history encoder is a causal transformer that maps each longitudinal
trajectory to time-indexed history representations. It uses:
\begin{itemize}[leftmargin=1.5em]
    \item $4$ layers of pre-norm self-attention;
    \item model dimension $256$, $8$ attention heads, and feed-forward dimension
    $1024$;
    \item sinusoidal temporal positional encodings up to the maximum sequence
    length;
    \item a causal attention mask preventing each time point from attending to
    future positions;
    \item zero initialization of selected residual output projections for stable
    training from scratch.
\end{itemize}
For a query current time $r$, the representation $h_r$ is extracted at position
$r$. For a support anchor label $Y_s$, the corresponding history representation
is extracted at $s-1$, so the support token represents the predictive mapping
from history and treatment through time $s-1$ to the outcome label at time $s$.

\subsection{PFN context encoder}
\label{app:pfn_layers}

The PFN context encoder performs in-context adaptation over support
trajectories. All $n_{\mathrm{ctx}}K_{\mathrm{sup}}$ support-anchor tokens and
the single query token attend bidirectionally to each other using full
self-attention, with a key-padding mask for padded support slots. No positional
encoding is added for the arbitrary order of support trajectories. Each PFN
layer uses multi-head attention, GELU feed-forward blocks, residual connections,
layer normalization, and zero-initialized residual output projections.

For support trajectory $j$ and anchor time $s_{jk}$, the support token is
\begin{equation}
    z^{\mathrm{ctx}}_{jk}
    =
    W_{\mathrm{tok}}\left[
        h_{j,s_{jk}-1}
        +W_xV_{j,s_{jk}-1}
        +W_CC_j
        +W_Gg(\calC),\;
        W_y(y_{j,s_{jk}},0)
    \right],
    \label{eq:support_token}
\end{equation}
where $g(\calC)$ contains symmetric support-level outcome statistics, including
the mean and standard deviation of support anchor outcomes. The query token at
current time $r$ is
\begin{equation}
    z^{\mathrm{qry}}_{r}
    =
    W_{\mathrm{tok}}\left[
        h^q_r
        +W_xV^q_r
        +W_CC_q
        +W_Gg(\calC),\;
        e_{\mathrm{qry}}
    \right],
    \label{eq:query_token}
\end{equation}
where $e_{\mathrm{qry}}$ is a learned query-label embedding. Thus, support
tokens contain observed anchor labels, whereas the query token marks the
unknown target to be predicted.

\subsection{Gaussian mixture head}
\label{app:gmm}

The final query representation $u_r$ is mapped to the parameters of a
five-component Gaussian mixture. The mixture weights, residual means, and scales
are computed as
\begin{align}
    \log\pi_r
    &=
    \log\operatorname{softmax}(W_\pi u_r/T_\pi),
    \qquad T_\pi=1.0,\\
    \Delta\mu_r
    &=
    7\tanh(W_\mu u_r/7),\\
    \sigma_r
    &=
    \clip_{[0.02,2.0]}
    \{\operatorname{softplus}(W_\sigma u_r)+0.02\}.
\end{align}
The component means are residualized around the most recent visible or
self-predicted query outcome:
\begin{equation}
    \mu_{r,k}
    =
    \clip_{[-12,12]}(y^q_r+\Delta\mu_{r,k}).
\end{equation}
The mean projection is initialized at zero, producing an initial persistence
predictor. This initialization stabilizes early training because
$\mu_{r,k}\approx y^q_r$ before the model has learned task-specific dynamics.

\subsection{Architecture and prior hyperparameters}
\label{app:hyperparams}

\begin{table}[h]
\centering
\caption{Architecture and synthetic-prior hyperparameters. Training and
optimization settings are reported separately in
Table~\ref{tab:optim_hparams}.}
\label{tab:hyperparams}
\small
\begin{tabular}{lll}
\toprule
Hyperparameter & Symbol & Value \\
\midrule
\multicolumn{3}{l}{\emph{Architecture}} \\
Model dimension & $d_{\mathrm{model}}$ & $256$ \\
Attention heads & $h$ & $8$ \\
History encoder layers & $N_{\mathrm{enc}}$ & $4$ \\
PFN layers & $N_{\mathrm{pfn}}$ & $6$ \\
feed-forward dimension & $d_{\mathrm{ff}}$ & $1024$ \\
Dropout & -- & $0.10$ \\
GMM components & $K_{\mathrm{gmm}}$ & $5$ \\
Mixture temperature & $T_\pi$ & $1.0$ \\
Minimum / maximum GMM std. dev. & -- & $0.02$ / $2.0$ \\
Residual mean bound before clipping & -- & $[-7,7]$ \\
Final mean clipping & -- & $[-12,12]$ \\
\midrule
\multicolumn{3}{l}{\emph{Synthetic prior}} \\
Maximum state dimension & $d_{S,\max}$ & $10$ \\
Observation window & $t_{\mathrm{obs}}$ & $1$--$60$ \\
Prediction horizon & $\tau$ & $1$--$5$ \\
Support size & $n_{\mathrm{ctx}}$ & $3$--$500$ \\
Number of treatments & $|\calA|$ & $4$ \\
Latent unit dimension & -- & $3$ \\
Support anchors & $K_{\mathrm{sup}}$ & $4$ \\
\bottomrule
\end{tabular}
\end{table}

\paragraph{Initialization.}
The self-attention output projections and final feed-forward projections in the
history and PFN transformer blocks are initialized at zero. The final GMM mean
projection is also initialized at zero, producing an initial persistence
forecast $\mu_{r,k}\approx y_r$.

\section{Loss function details}
\label{app:loss}

The model is trained with a Gaussian-mixture one-step predictive loss. The total
loss is
\begin{equation}
    \mathcal{L}
    =
    \mathcal{L}_{\mathrm{NLL}}
    +\lambda_m\mathcal{L}_{\mathrm{mean}}
    +\lambda_c\mathcal{L}_{\mathrm{conc}},
    \qquad
    \lambda_m=0.25,\quad
    \lambda_c=0.03.
\end{equation}

\paragraph{Robust NLL.}
For normalized target $z$, the Gaussian-mixture negative log-likelihood is
\begin{equation}
    \ell_{\mathrm{NLL}}
    =
    -\log
    \sum_{k=1}^{K_{\mathrm{gmm}}}
    \pi_k
    \frac{1}{\sigma_k\sqrt{2\pi}}
    \exp\left\{
        -\frac{1}{2}
        \left(\frac{z-\mu_k}{\sigma_k}\right)^2
    \right\}.
\end{equation}
The implementation uses a linear tail for very large NLL values:
\begin{equation}
    \tilde\ell_{\mathrm{NLL}}
    =
    \begin{cases}
        \ell_{\mathrm{NLL}}, & \ell_{\mathrm{NLL}}\le 15,\\
        15+0.01(\ell_{\mathrm{NLL}}-15), & \ell_{\mathrm{NLL}}>15.
    \end{cases}
\end{equation}
This robustification prevents rare unstable synthetic examples from dominating
gradients while retaining a loss signal.

\paragraph{Mean loss.}
The predictive mean is
\begin{equation}
    \hat z=\sum_k\pi_k\mu_k.
\end{equation}
The auxiliary mean loss is
\begin{equation}
    \mathcal{L}_{\mathrm{mean}}
    =
    \operatorname{Huber}_{\delta=3}(\hat z-z).
\end{equation}
This term provides a direct gradient signal for point prediction and stabilizes
early optimization.

\paragraph{Concentration penalty.}
The mixture-concentration penalty is
\begin{equation}
    \mathcal{L}_{\mathrm{conc}}
    =
    \left[\max_k\pi_k-0.90\right]_+.
\end{equation}
It discourages premature collapse of the mixture distribution onto a single
component.

\section{Training procedure and stability}
\label{app:training}

\paragraph{Optimizer and schedule.}
The model is optimized with AdamW \citep{loshchilov2019decoupled}. Weight decay
is applied to ordinary dense weight matrices, while biases, layer-normalization
parameters, query embeddings, context-statistic encoders, and static-feature
encoders are excluded from weight decay. The learning rate is warmed up
linearly for $400$ optimizer steps and then cosine-decayed to $2\%$ of the peak
over $10{,}000$ steps.

\paragraph{Stochastic PFN depth.}
At each training step, the number of active PFN layers is sampled uniformly from
$\{3,\ldots,6\}$. This stochastic-depth-like regularization encourages useful
intermediate-depth representations and reduces dependence on the deepest PFN
stack.

\paragraph{Gradient accumulation and clipping.}
Gradients are accumulated over $16$ micro-batches, giving an effective batch
size of $256$ synthetic episodes. Before each optimizer step, gradients are
unscaled under automatic mixed precision and clipped using a threshold that
increases linearly from $0.5$ to $1.5$ over the first $4000$ optimizer steps.
This applies tighter clipping during early training and looser clipping after
the model stabilizes.

\paragraph{Numerical stability safeguards.}
Training includes several numerical safeguards:
\begin{enumerate}[leftmargin=1.5em]
    \item If the loss is non-finite, the batch is skipped and the AMP loss scale
    is reduced.
    \item If the global gradient norm is non-finite, the optimizer step is
    skipped.
    \item The GMM loss computation is upcast to FP32 before log-sum-exp
    operations.
    \item Normalized inputs and targets are clipped, and GMM standard deviations
    are bounded in $[0.02,2.0]$.
\end{enumerate}

\begin{table}[h]
\centering
\caption{Training and optimization hyperparameters. Gradient accumulation gives
an effective batch size of $256$ synthetic episodes.}
\label{tab:optim_hparams}
\begin{tabular}{ll}
\toprule
Quantity & Value \\
\midrule
Batch size & $16$ \\
Gradient accumulation & $16$ steps \\
Effective batch size & $256$ \\
Optimizer & AdamW \\
Learning rate & $3\times10^{-4}$ \\
Weight decay & $10^{-5}$, excluding bias/norm/static/query parameters \\
Warmup & $400$ optimizer steps \\
Schedule & Cosine decay to $0.02$ of base LR over $10{,}000$ steps \\
Maximum optimizer steps & $10{,}000$ \\
Random PFN depth during training & Uniformly $3$--$6$ layers \\
Gradient clipping & Linear ramp $0.5$ to $1.5$ over $4000$ steps \\
Checkpoint interval & $500$ optimizer steps \\
Mixed precision & Enabled on CUDA \\
Seed & $42$ \\
\bottomrule
\end{tabular}
\end{table}

\section{Autoregressive rollout}
\label{app:rollout}

\begin{algorithm}[h]
\caption{Plug-in autoregressive counterfactual rollout}
\label{alg:rollout}
\begin{algorithmic}[1]
\State \textbf{Input:} frozen model $q_\theta$, support trajectories $\calC$,
query history through $t_{\mathrm{obs}}$, future treatment sequence
$a_{t_{\mathrm{obs}}:t_\star-1}$.
\State Initialize a writable query sequence $\widetilde X^q$ with future query
covariates hidden and future query outcomes hidden.
\For{$r=t_{\mathrm{obs}},\ldots,t_\star-1$}
    \State Run one-step inference to obtain mixture
    $\{\pi_{r,k},\mu_{r,k},\sigma_{r,k}\}_{k=1}^{5}$.
    \State Compute the plug-in mean
    $\hat y_{r+1}=\sum_k\pi_{r,k}\mu_{r,k}$.
    \If{$r+1$ is within the input sequence}
        \State Insert $\hat y_{r+1}$ into the query outcome channel at time
        $r+1$.
    \EndIf
    \State Keep future query covariates at the sentinel value.
\EndFor
\State \Return final mixture and mean prediction at $t_\star$.
\end{algorithmic}
\end{algorithm}

The final mixture is conditional on the self-fed mean trajectory. Thus the
reported distribution does not integrate over all possible intermediate outcome
paths. A stochastic ancestral variant could sample from the mixture at each
intermediate step and repeat the rollout to approximate full path-level
uncertainty.

\section{Evaluation datasets}
\label{app:evaluation_datasets}

We evaluate on four longitudinal treatment-response benchmarks: a cancer tumor
growth simulator, a semi-mechanistic warfarin pharmacokinetic/pharmacodynamic
(PK/PD) simulator, an HIV ODE simulator based on Adams/WhyNot dynamics, and a
factual MIMIC-III ICU rolling-origin benchmark. Cancer, warfarin, and HIV are
branchable simulated or semi-mechanistic systems: for these domains,
counterfactual outcomes under alternative future treatment sequences are
available by replaying the same patient-specific dynamics under intervened
treatments. MIMIC-III contains real observational ICU data and does not reveal
outcomes under unobserved interventions; it is therefore used only for factual
rolling-origin prediction under observed future treatments. This distinction
follows prior longitudinal counterfactual evaluations, where simulated systems
provide ground-truth counterfactual labels and real ICU cohorts provide factual
temporal prediction benchmarks
\citep{lim2018rmsn,bica2020crn,melnychuk2022causaltransformer,li2021gnet}.

\paragraph{Common task construction.}
Let $i$ index patients and let $t$ index discrete time. We write $S_{i,t}$ for time-varying covariates or simulator state variables,
$A_{i,t}$ for treatment, $Y_{i,t}$ for the scalar target outcome, and $C_i$ for
time-invariant patient features. The model-facing state is
$X_{i,t}=(S_{i,t},Y_{i,t})$ when the outcome channel is included. Each rolling-origin query observes a patient history up to an origin
time and asks for either the next outcome or the outcome after a supplied future
treatment sequence.

Across domains, raw trajectories have length $T=60$, the projection horizon is
$H=5$, and the configured minimum observed history length is $t_{\min}=10$.
For each domain, tasks vary the confounding level $\gamma$, support size
$n_{\mathrm{sup}}\in\{40,80,160,320,500\}$, and random repetition. In the
simulated and semi-mechanistic domains, $\gamma$ controls how strongly the
behavior policy depends on current patient state and therefore controls the
strength of time-varying confounding. In MIMIC-III, $\gamma$ is retained only as
a stratification variable for consistent task organization and does not modify
the observed data.

All reported metrics use support-only normalization. The rolling-origin filters,
indexing conventions, and clipping rules used for scoring are given in
Appendix~\ref{app:evaluation_protocol_details}. The domain-level summary is
shown in Table~\ref{tab:evaluation_dataset_summary}.

\paragraph{Counterfactual and factual test queries.}
For cancer, warfarin, and HIV, one-step query rows branch from a factual patient
state and evaluate alternative treatments. Horizon-5 query rows branch from a
factual origin and replay the same patient-specific dynamics forward under
randomly sampled treatment sequences of length $H=5$. For MIMIC-III, both
one-step and horizon-5 rows are factual rolling-origin predictions: future
treatments are the observed ICU treatments, not interventions. In all domains,
scored rows follow the common evaluation protocol in
Appendix~\ref{app:evaluation_protocol_details}.

\begin{table}[t]
\centering
\small
\caption{Evaluation datasets. Simulated and semi-mechanistic domains provide
branchable counterfactual labels; MIMIC-III is factual-only and evaluates
temporal prediction under observed clinical practice.}
\label{tab:evaluation_dataset_summary}
\begin{tabular}{p{0.16\linewidth} p{0.18\linewidth} p{0.25\linewidth} p{0.18\linewidth} p{0.17\linewidth}}
\toprule
Domain & Data type & Treatments & Target & Query construction \\
\midrule
Cancer tumor growth &
Fully simulated PK/PD tumor dynamics &
Four discrete treatment actions induced by chemotherapy/radiotherapy combinations &
$\log(1+\text{clipped tumor volume})$ &
One-step: all four joint actions. Multi-step: random treatment sequences of
length $H$. \\
\midrule
Warfarin &
Semi-mechanistic PK/PD simulator &
Four dose classes corresponding to $0,2,5,10$ mg/day, delivered in each 4-hour bin &
INR &
One-step: all four dose actions. Multi-step: random dose sequences of length
$H$. \\
\midrule
HIV &
Adams/WhyNot-style six-compartment ODE simulator &
Four antiretroviral regimens: none, PI only, RTI only, RTI+PI &
$\log_{10}(1+\text{free virus})$ &
One-step: all four regimens. Multi-step: random regimen sequences of length
$H$. \\
\midrule
MIMIC-III &
Real factual ICU time series from MIMIC-Extract &
Four observed treatment classes from vasopressors and ventilation:
none, vaso, vent, vaso+vent &
Diastolic blood pressure &
Factual-only rolling-origin rows; future treatments are observed ICU treatments,
not interventions. \\
\bottomrule
\end{tabular}
\end{table}

All targets are normalized only at scoring time using support-set statistics, as
described in Appendix~\ref{app:evaluation_protocol_details}.

\subsection{Cancer tumor growth simulator}
\label{app:data_cancer}

\paragraph{Background.}
The cancer benchmark follows the tumor-growth simulator used in RMSN, CRN,
Causal Transformer, G-Net, and related longitudinal counterfactual evaluations
\citep{lim2018rmsn,bica2020crn,melnychuk2022causaltransformer,li2021gnet,geng2017prediction}.
The simulator represents non-small-cell lung cancer tumor volume evolving under
chemotherapy and radiotherapy. It combines Gompertz-style tumor growth with
linear-quadratic radiotherapy effects and log-cell-kill chemotherapy effects.
Treatment assignment depends on recent tumor history, producing time-varying
confounding.

\paragraph{Dynamics.}
Let $Y_{i,t}^{\mathrm{raw}}$ denote raw tumor volume. Diameter and volume are
converted by
\begin{equation}
    \operatorname{Vol}(d)
    =
    \frac{4}{3}\pi\left(\frac{d}{2}\right)^3,
    \qquad
    \operatorname{Diam}(y)
    =
    2\left(\frac{y}{4\pi/3}\right)^{1/3}.
\end{equation}
The carrying capacity is $K=\operatorname{Vol}(30)$, and the death threshold is
$Y_{\max}=\operatorname{Vol}(13)$. Patient-specific parameters include tumor
growth rate $\rho_i$, radiosensitivity coefficients $\alpha_i$ and
$\beta_i=\alpha_i/10$, and chemotherapy kill coefficient $\beta_i^c$.

At each time $t$, chemotherapy and radiotherapy are assigned by Bernoulli
policies depending on recent tumor diameter:
\begin{equation}
    \bar D_{i,t}^{(15)}
    =
    \frac{1}{|\mathcal W_t|}
    \sum_{s\in\mathcal W_t}
    \operatorname{Diam}\!\left(Y_{i,s}^{\mathrm{raw}}\right),
    \qquad
    \mathcal W_t=\{\max(0,t-15),\ldots,t\}.
\end{equation}
The behavior policy is
\begin{equation}
    \Pr(A^c_{i,t}=1\mid \bar H_{i,t})
    =
    \Pr(A^r_{i,t}=1\mid \bar H_{i,t})
    =
    \sigma\!\left[
        \frac{\gamma}{D_{\max}}
        \left(
            \bar D_{i,t}^{(15)}-\frac{D_{\max}}{2}
        \right)
    \right],
\end{equation}
where $A^c_{i,t}$ and $A^r_{i,t}$ denote chemotherapy and radiotherapy
indicators and $D_{\max}=13$. Larger $\gamma$ strengthens the dependence of
treatment assignment on tumor history.

Chemotherapy is administered as a dose of 5 units when $A^c_{i,t}=1$, with
half-life one time step:
\begin{equation}
    C_{i,t}=2^{-1}C_{i,t-1}+5A^c_{i,t}.
\end{equation}
Radiotherapy is an immediate dose $R_{i,t}=2A^r_{i,t}$. Raw tumor volume evolves
as
\begin{equation}
\begin{aligned}
    Y_{i,t+1}^{\mathrm{raw}}
    =
    Y_{i,t}^{\mathrm{raw}}
    \bigg[
        1
        +\rho_i\log\!\left(\frac{K}{Y_{i,t}^{\mathrm{raw}}}\right)
        -\beta^c_i C_{i,t}
        -\left(\alpha_i R_{i,t}+\beta_i R_{i,t}^2\right)
        +\epsilon_{i,t}
    \bigg],
    \qquad
    \epsilon_{i,t}\sim\mathcal N(0,0.01^2).
\end{aligned}
\label{eq:cancer_dyn}
\end{equation}

\paragraph{Outcome representation.}
The model-facing cancer outcome is clipped log-volume,
\begin{equation}
    Y_{i,t}
    =
    \log\!\left(1+\min\{Y_{i,t}^{\mathrm{raw}},Y_{\max}\}\right).
\end{equation}
Support and query outcomes are normalized from this transformed scale.

\subsection{Warfarin semi-mechanistic PK/PD simulator}
\label{app:data_warfarin}

\paragraph{Background.}
The warfarin benchmark is a semi-mechanistic PK/PD simulator motivated by
standard warfarin dose--response models: oral absorption and elimination,
delayed anticoagulant response through inhibition of vitamin-K-dependent
coagulation-factor synthesis, INR readout from clotting-factor activity, and
patient heterogeneity driven by CYP2C9 metabolism, VKORC1 sensitivity, age,
dietary vitamin K, and adherence \citep{holford1986clinical,hamberg2010pkpd,iwpc2009estimation}.

\paragraph{PK model.}
Each time step is a 4-hour bin. The treatment space is
$\mathcal A=\{0,1,2,3\}$, corresponding to daily dose levels
\begin{equation}
    (0,2,5,10)\ \mathrm{mg/day}.
\end{equation}
The PK state follows a gut $\rightarrow$ plasma $\rightarrow$ effect-site model:
\begin{align}
    \dot A_g &= -k_a A_g,\\
    \dot C_p &= k_a A_g/V_d - k_e C_p,\\
    \dot C_e &= k_{e0}(C_p-C_e),
\end{align}
where $A_g$ is gut depot amount, $C_p$ is plasma concentration, $C_e$ is
effect-site concentration, $k_a$ is absorption, $k_e=\mathrm{CL}/V_d$ is
elimination, and $k_{e0}$ is effect-site equilibration.

\paragraph{PD model.}
Effect-site concentration inhibits vitamin-K-dependent synthesis through an
$E_{\max}$ model:
\begin{equation}
    I(t)
    =
    E_{\max}\frac{C_e(t)^h}{C_e(t)^h+\mathrm{EC}_{50}^h}.
\end{equation}
Each coagulation factor
$f\in\{\mathrm{II},\mathrm{VII},\mathrm{X},\mathrm{PC}\}$ follows delayed
turnover:
\begin{equation}
    \dot f(t)
    =
    k_{\mathrm{out},f}
    \left[
        \mathrm{VK}(t)\{1-s_f I(t)\}-f(t)
    \right].
\end{equation}
INR is computed from factor deficits:
\begin{align}
    \Delta_f(t)
    &=
    1.30[1-f_{\mathrm{VII}}(t)]_+
    +0.95[1-f_{\mathrm{X}}(t)]_+
    +0.80[1-f_{\mathrm{II}}(t)]_+,\\
    \mathrm{PT}(t)
    &=
    1+1.6\Delta_f(t)+0.70\Delta_f(t)^2,\\
    \mathrm{INR}(t)
    &=
    \mathrm{INR}_0\,\mathrm{PT}(t)^{\mathrm{ISI}}.
\end{align}

\paragraph{Patient heterogeneity and confounding.}
Patient heterogeneity includes CYP metabolism class, VKORC1 sensitivity, age,
clearance, absorption, effect-site kinetics, pharmacodynamic sensitivity,
vitamin-K baseline, clinic bias, and maintenance-dose requirement. The behavior
policy is a softmax over dose classes whose logits depend on current INR,
distance from the therapeutic range, INR trend, effect-site concentration,
recent dose load, adherence, age, maintenance-dose class, and clinic bias. The
confounding parameter $\gamma$ scales the INR-dependent policy terms, increasing
the dependence of dosing on current patient state.

\paragraph{State and outcome.}
The visible model-facing state is 10-dimensional:
\begin{equation}
\begin{aligned}
X_t =
\big(
&C^{\mathrm{plasma}}_t,\,
C^{\mathrm{effect}}_t,\,
F^{\mathrm{II}}_t,\,
F^{\mathrm{VII}}_t,\,
K^{\mathrm{vit}}_t,\,
\mathrm{INR}_t,\\
&\mathrm{doseLoad}_{7d,t},\,
\mathrm{CYP}_i,\,
\mathrm{VKORC1}_i,\,
\mathrm{ageNorm}_i
\big).
\end{aligned}
\end{equation}
The scalar target is
\begin{equation}
    Y_t=\mathrm{INR}_t.
\end{equation}

\subsection{HIV Adams/WhyNot ODE simulator}
\label{app:data_hiv}

\paragraph{Background.}
The HIV benchmark is based on the six-compartment Adams HIV treatment ODE used
in the WhyNot simulator suite \citep{adams2004dynamic,miller2020whynot}. It
models immunological and virological dynamics under antiretroviral therapy and
allows patient-specific counterfactual evaluation by intervening on future
treatment regimens.

\paragraph{ODE dynamics.}
The raw state is
\begin{equation}
    S_t=(T_{1,t},T^*_{1,t},T_{2,t},T^*_{2,t},V_t,E_t),
\end{equation}
where $T_1,T_2$ are uninfected target-cell populations, $T^*_1,T^*_2$ are
infected cell populations, $V$ is free virus, and $E$ is immune response. The
dynamics follow
\begin{align}
  \dot T_1 &= \lambda_1-d_1T_1-(1-\epsilon_1)k_1VT_1,\\
  \dot T_1^* &= (1-\epsilon_1)k_1VT_1-\delta T_1^*-m_1ET_1^*,\\
  \dot T_2 &= \lambda_2-d_2T_2-(1-f\epsilon_1)k_2VT_2,\\
  \dot T_2^* &= (1-f\epsilon_1)k_2VT_2-\delta T_2^*-m_2ET_2^*,\\
  \dot V &=
  (1-\epsilon_2)N_T\delta(T_1^*+T_2^*)-cV
  -\left[
      (1-\epsilon_1)\rho_1k_1T_1
      +(1-f\epsilon_1)\rho_2k_2T_2
   \right]V,\\
  \dot E &=
  \lambda_E
  +\frac{b_E(T_1^*+T_2^*)}{T_1^*+T_2^*+K_B}E
  -\frac{d_E(T_1^*+T_2^*)}{T_1^*+T_2^*+K_D}E
  -\delta_EE.
\end{align}

\paragraph{Treatment space.}
There are four treatment regimens:
\begin{equation}
\begin{array}{ccl}
0 &:& \text{no therapy},\quad (\epsilon_1,\epsilon_2)=(0,0),\\
1 &:& \text{PI only},\quad (\epsilon_1,\epsilon_2)=(0,0.3),\\
2 &:& \text{RTI only},\quad (\epsilon_1,\epsilon_2)=(0.7,0),\\
3 &:& \text{RTI+PI},\quad (\epsilon_1,\epsilon_2)=(0.7,0.3).
\end{array}
\end{equation}

\paragraph{Patient heterogeneity and confounding.}
Patient heterogeneity is introduced through perturbations of the ODE parameters,
individual RTI/PI efficacy scales, viral and immune thresholds, and policy
aggressiveness. The behavior policy computes a severity score from current
$\log_{10}(V+1)$ and $\log_{10}(E+1)$, scales this score by $\gamma$, adds
inertia for the previous treatment, and samples one of the four regimens from a
softmax. Larger $\gamma$ increases dependence of treatment choice on biological
state.

\paragraph{State and outcome.}
The model-facing state is the log-transformed compartment vector
\begin{equation}
    X_{t,k}=\log_{10}(1+S_{t,k}),\qquad k=1,\ldots,6.
\end{equation}
The scalar target is transformed free virus,
\begin{equation}
    Y_t=\log_{10}(1+V_t).
\end{equation}

\subsection{MIMIC-III factual ICU rolling-origin benchmark}
\label{app:data_mimic}

\paragraph{Background.}
The MIMIC benchmark is constructed from MIMIC-III ICU stays using a
MIMIC-Extract-style hourly representation \citep{PhysioNet-mimiciii-1.4,johnson2016mimic,goldberger2000physionet,wang2020mimicextract,harutyunyan2019mimicbenchmarks}.
Because MIMIC-III does not provide ground-truth counterfactual outcomes, we use
it only for factual rolling-origin prediction under observed future treatments.

\paragraph{State and preprocessing.}
Each ICU stay is treated as a single hourly sequence. The model-facing state is
10-dimensional:
\begin{equation}
\begin{aligned}
X_t = (&
\text{diastolic blood pressure},
\text{mean blood pressure},
\text{oxygen saturation},
\text{heart rate},
\text{respiratory rate},\\
&
\text{Glasgow Coma Scale total},
\text{glucose},
\text{creatinine},
\text{bicarbonate},
\text{sodium}
).
\end{aligned}
\end{equation}
Static features are derived from demographic variables and represented by a
fixed-dimensional vector.

\paragraph{Treatment space and outcome.}
The two binary treatment indicators are vasopressor administration
$\mathrm{vaso}_t$ and mechanical ventilation $\mathrm{vent}_t$. They are
combined into the four-valued treatment used by the shared model interface,
\begin{equation}
    A_t
    =
    \mathbb{I}\{\mathrm{vaso}_t\}
    +2\mathbb{I}\{\mathrm{vent}_t\},
\end{equation}
with mapping
\begin{equation}
    0:\text{none},\qquad
    1:\text{vaso},\qquad
    2:\text{vent},\qquad
    3:\text{vaso+vent}.
\end{equation}
The scalar target is
\begin{equation}
    Y_t=\text{diastolic blood pressure}_t.
\end{equation}
MIMIC-III results should be interpreted as factual temporal prediction results,
not as validation of counterfactual treatment effects.

\section{Baseline models}
\label{app:baseline_models}

We compare against six longitudinal baselines: a classical Marginal Structural
Model (MSM), Recurrent Marginal Structural Networks (RMSN), G-Net,
Counterfactual Recurrent Networks (CRN), Causal Transformer (CT), and
G-Transformer (GT). Together, these baselines cover the main adjustment
strategies used in longitudinal treatment-response prediction:
inverse-probability weighting \citep{robins2000msm}, recurrent marginal
structural modeling \citep{lim2018rmsn}, neural $g$-computation
\citep{li2021gnet}, adversarial representation balancing \citep{bica2020crn},
and transformer-based counterfactual sequence modeling
\citep{melnychuk2022causaltransformer,xiong2024gtransformer}.

\paragraph{Common notation.}
For unit $i$ and
time $t$, let $x_{i,t}\in\mathbb{R}^{d_x}$ denote the baseline-model covariate input corresponding to the time-varying covariates $S_{i,t}$, let
$y_{i,t}\in\mathbb{R}$ denote the scalar outcome $Y_{i,t}$, let
$a_{i,t}\in\{0,1,2,3\}$ denote the treatment $A_{i,t}$, and let
$c_i\in\mathbb{R}^{5}$ denote static covariates corresponding to $C_i$. We write
$\tilde y_{i,t}$ for the normalized outcome used by the baseline training code.
All baselines follow the same inclusive observation-time convention as
Section~\ref{sec:problem_formulation}: the model observes the history through
$t_{\mathrm{obs}}$, receives the first planned treatment
$a_{i,t_{\mathrm{obs}}}$, and predicts future outcomes under the supplied
treatment sequence.

\paragraph{Normalization and metrics.}
Continuous inputs are normalized using statistics computed from the support
trajectories of the corresponding benchmark file. The primary reported metric is
normalized RMSE under the shared support-only evaluation normalization. For
\CausalLongPFN{}, predictions are first converted from the model's internal
PFN-context normalization to raw outcome units and then to the shared evaluation
normalization. Full scoring details, including clipping rules, are given in
Appendix~\ref{app:evaluation_protocol_details}.

\paragraph{Treatment encodings.}
All methods use the same four-valued treatment space
$a_t\in\{0,1,2,3\}$. For models that require vector-valued treatment inputs, we
use either the one-hot encoding
\begin{equation}
    \phi_4(a_t)=e_{a_t}\in\{0,1\}^{4},
\end{equation}
or the equivalent two-bit decomposition
\begin{equation}
    \phi_2(a_t)
    =
    \left(a_t \bmod 2,\; \left\lfloor a_t/2\right\rfloor\right)
    \in\{0,1\}^{2}.
\end{equation}
\CausalLongPFN{} and the neural baselines use the four-valued treatment input,
whereas MSM and RMSN use $\phi_2$ in their propensity-weighting components.

\paragraph{Hyperparameter tuning protocol.}
Baseline hyperparameters are selected using only support trajectories from the
target domain. For each baseline, the evaluation runner performs an initial
random search over the method-specific search space in
Table~\ref{tab:baseline_hparam_search_spaces} for the first dataset in each
$(\text{domain}, n_{\mathrm{sup}})$ tuning group. Candidate configurations are
trained on a support-training split and ranked using normalized RMSE on a
held-out support-validation split. The top cached candidate is then reused and
re-evaluated on subsequent support-validation splits within the same tuning
group. The selected configuration is finally refit on the full support set
before query evaluation. Query outcomes are never used for hyperparameter
selection.

This protocol gives all baselines domain-specific supervision and
validation-based model selection. In contrast, \CausalLongPFN{} is evaluated as
a frozen pretrained model: no target-domain gradients are taken, no validation
set is used for model selection, and no target-domain hyperparameters are tuned.

\begin{table}[t]
\centering
\scriptsize
\setlength{\tabcolsep}{3pt}
\caption{Baseline hyperparameter search spaces used by the evaluation runner.
Each baseline is tuned using support-set validation only, with random search
over the listed discrete candidates. \CausalLongPFN{} is not included because it
is evaluated as a frozen model without target-domain tuning.}
\label{tab:baseline_hparam_search_spaces}
\begin{tabular}{p{0.12\linewidth} p{0.42\linewidth} p{0.38\linewidth}}
\toprule
Method & Architecture / model search space & Optimization / regularization search space \\
\midrule
MSM &
Regressor $\in\{\mathrm{linear},\mathrm{ridge}\}$;
lag features $=2$;
ridge penalty $\alpha\in\{0.1,1.0,10.0\}$ &
Stabilized treatment weights are clipped at support-set quantiles
$(0.01,0.99)$. Logistic propensity models use a maximum iteration count $500$ and
$C=10^6$. \\
\midrule
RMSN &
Number of recurrent layers $\in\{1,2\}$.
Encoder and decoder hidden widths are selected from a data-dimensionality-aware
grid. Let $d_x$ be the number of time-varying covariates,
$C_{\mathrm{hist}}=d_x+1+2+5$, and $C_{\mathrm{dec}}=1+2+5$. Candidate widths
are obtained by multiplying $C_{\mathrm{hist}}$ and $C_{\mathrm{dec}}$ by
$\{0.5,1,2,4\}$, rounding to a multiple of $16$, clipping to $[32,160]$, and
unioning with $\{32,48,64,96,128\}$. &
Dropout $\in\{0.1,0.2,0.3,0.4,0.5\}$;
propensity learning rate $\in\{10^{-2},10^{-3},10^{-4}\}$;
encoder learning rate $\in\{10^{-2},10^{-3},10^{-4}\}$;
decoder learning rate $\in\{10^{-2},10^{-3},10^{-4}\}$;
encoder batch size $\in\{64,128,256\}$;
decoder batch size $\in\{256,512,1024\}$;
gradient clipping $\in\{0.5,1.0,2.0,4.0\}$. \\
\midrule
G-Net &
Hidden size $\in\{48,64,96,128\}$;
representation size $\in\{48,64,96\}$;
number of recurrent layers $\in\{1,2\}$. &
Dropout $\in\{0.05,0.10,0.20\}$;
learning rate $\in\{10^{-3},3{\times}10^{-4}\}$;
batch size $\in\{32,64\}$;
epochs $\in\{80,120\}$;
covariate/vitals loss weight $\in\{0.15,0.25,0.35\}$. \\
\midrule
CRN &
Number of recurrent layers $\in\{1,2\}$.
Hidden, balanced-representation, and fully connected widths are selected from a
data-dimensionality-aware grid. Let
$C_{\mathrm{hist}}=4+d_x+1+5$ and $C_{\mathrm{dec}}=4+1+5$. Width candidates
are obtained by multiplying these quantities by $\{0.5,1,2,4\}$, rounding to a
multiple of $16$, clipping to $[32,160]$, and unioning hidden and balanced
widths with $\{32,48,64,96,128\}$ and fully connected widths with
$\{32,64,96,128,192\}$. &
Dropout $\in\{0.1,0.2,0.3,0.4,0.5\}$;
encoder learning rate $\in\{10^{-2},10^{-3},10^{-4}\}$;
decoder learning rate $\in\{10^{-2},10^{-3},10^{-4}\}$;
encoder batch size $\in\{64,128,256\}$;
decoder batch size $\in\{256,512,1024\}$;
gradient clipping $\in\{0.5,1.0,2.0\}$. \\
\midrule
Causal Transformer &
Transformer layers $\in\{2,3,4\}$;
attention heads $\in\{2,4\}$;
sequence hidden size $\in\{64,96,128\}$;
balanced-representation size $\in\{64,96,128\}$;
fully connected hidden size $\in\{64,96,128\}$. &
Dropout $\in\{0.05,0.10,0.20\}$;
learning rate $\in\{10^{-3},3{\times}10^{-4},10^{-4}\}$;
weight decay $\in\{10^{-5},10^{-4},10^{-3}\}$;
batch size $\in\{32,64\}$;
gradient clipping $\in\{0.5,1.0\}$;
treatment loss weight $\in\{0.05,0.10,0.20\}$. \\
\midrule
G-Transformer &
Transformer layers $\in\{2,3,4\}$;
attention heads $\in\{2,4\}$;
model dimension $\in\{32,48,64,96,128\}$;
balanced-representation size $\in\{32,48,64,96,128\}$;
fully connected hidden size $\in\{32,64,96,128,192\}$. &
Dropout $\in\{0.1,0.2,0.3\}$;
learning rate $\in\{10^{-3},10^{-4},10^{-5}\}$;
weight decay $\in\{10^{-5},10^{-4},10^{-3}\}$;
batch size $\in\{16,32,64\}$. \\
\bottomrule
\end{tabular}
\end{table}

\subsection{Marginal Structural Model}
\label{app:baseline_msm}

The MSM baseline adjusts for time-varying confounding through stabilized inverse
probability of treatment weights \citep{robins2000msm}. The numerator
propensity model uses prior treatment history,
\begin{equation}
    z^{\mathrm{num}}_{i,t}
    =
    \sum_{u=0}^{t-1}\phi_2(a_{i,u}),
\end{equation}
whereas the denominator propensity model conditions on treatment, covariate,
outcome, and static history,
\begin{equation}
    z^{\mathrm{den}}_{i,t}
    =
    \left[
    \sum_{u=0}^{t-1}\phi_2(a_{i,u}),\;
    x_{i,t-L_{\mathrm{lag}}:t},\;
    \tilde y_{i,t-L_{\mathrm{lag}}:t},\;
    c_i
    \right].
\end{equation}
The stabilized treatment ratio at time $t$ is
\begin{equation}
    w_{i,t}
    =
    \frac{
    \prod_{b=1}^{2}
    p_{\mathrm{num},b}
    \left(a_{i,t,b}\mid z^{\mathrm{num}}_{i,t}\right)
    }{
    \prod_{b=1}^{2}
    p_{\mathrm{den},b}
    \left(a_{i,t,b}\mid z^{\mathrm{den}}_{i,t}\right)
    },
\end{equation}
where $a_{i,t,b}$ is the $b$th treatment bit. The outcome model is direct in
horizon: for horizon $\tau$, it predicts $\tilde y_{i,t+\tau}$ from
\begin{equation}
    z^{\mathrm{MSM}}_{i,t,\tau}
    =
    \left[
    z^{\mathrm{den}}_{i,t},\;
    \sum_{u=t}^{t+\tau-1}\phi_2(a_{i,u})
    \right].
\end{equation}

\subsection{Recurrent Marginal Structural Networks}
\label{app:baseline_rmsn}

RMSN replaces the propensity and outcome regressions of MSM with recurrent
neural networks \citep{lim2018rmsn}. A treatment-only propensity network
predicts the current treatment from previous treatments,
\begin{equation}
    \hat p^{\mathrm{num}}_{i,t}
    =
    \sigma\!\left(f_{\mathrm{prop},T}(\phi_2(a_{i,0:t-1}))\right),
\end{equation}
while a history-dependent propensity network predicts treatment from previous
treatments, covariates, outcomes, and static features,
\begin{equation}
    \hat p^{\mathrm{den}}_{i,t}
    =
    \sigma\!\left(
    f_{\mathrm{prop},H}(\phi_2(a_{i,0:t-1}),x_{i,0:t},
    \tilde y_{i,0:t},c_i)
    \right).
\end{equation}
Stabilized weights are computed from the ratio of these probabilities and used
to train recurrent encoder and decoder outcome models. The encoder predicts
one-step outcomes from
\begin{equation}
    u^{\mathrm{enc}}_{i,t}
    =
    \left[x_{i,t},\;\tilde y_{i,t},\;\phi_2(a_{i,t}),\;c_i\right],
\end{equation}
and the decoder performs autoregressive multi-step rollout under the planned
future treatment sequence.

\subsection{G-Net}
\label{app:baseline_gnet}

G-Net is a recurrent neural $g$-computation baseline \citep{li2021gnet}. It
models the next outcome jointly with the next covariates and then rolls the
system forward under a planned treatment sequence. At time $t$, the input is
\begin{equation}
    u^{\mathrm{GNet}}_{i,t}
    =
    \left[\phi_4(a_{i,t}),\;x_{i,t},\;\tilde y_{i,t},\;c_i\right].
\end{equation}
The training objective combines next-outcome prediction with next-covariate
prediction:
\begin{equation}
    \mathcal{L}_{\mathrm{GNet}}
    =
    \frac{
    \sum_{i,t}m_{i,t}
    \left(\hat y_{i,t+1}-\tilde y_{i,t+1}\right)^2
    }{
    \sum_{i,t}m_{i,t}
    }
    +
    \lambda_x
    \frac{
    \sum_{i,t}m^{x}_{i,t}
    \left\|\hat x_{i,t+1}-x_{i,t+1}\right\|_2^2
    }{
    \sum_{i,t}m^{x}_{i,t}
    }.
\end{equation}
At test time, predicted outcomes and covariates are fed back autoregressively,
implementing plug-in neural $g$-computation under the supplied treatment
sequence.

\subsection{Counterfactual Recurrent Network}
\label{app:baseline_crn}

CRN learns balanced recurrent representations by combining outcome prediction
with adversarial treatment prediction \citep{bica2020crn}. The encoder input is
\begin{equation}
    u^{\mathrm{CRN}}_{i,t}
    =
    \left[\phi_4(a_{i,t-1}),\;x_{i,t},\;\tilde y_{i,t},\;c_i\right],
\end{equation}
with a zero previous-treatment vector at $t=0$. The recurrent representation is
mapped to a balanced representation
\begin{equation}
    b_{i,t}=\operatorname{ELU}(W_b h_{i,t}+c_b).
\end{equation}
The treatment head predicts $a_{i,t}$ from a gradient-reversed version of
$b_{i,t}$, while the outcome head predicts $\tilde y_{i,t+1}$ from
$[b_{i,t},\phi_4(a_{i,t})]$. The objective is
\begin{equation}
    \mathcal{L}_{\mathrm{CRN}}
    =
    \frac{\sum_{i,t}m_{i,t}
    \left(\hat y_{i,t+1}-\tilde y_{i,t+1}\right)^2}{\sum_{i,t}m_{i,t}}
    +
    \lambda_a\,
    \mathrm{CE}_{\mathrm{active}}(\hat a_{i,t},a_{i,t}).
\end{equation}
A recurrent decoder performs autoregressive rollout under planned treatments.

\subsection{Causal Transformer}
\label{app:baseline_ct}

Causal Transformer replaces the recurrent backbone of CRN with a multi-input
transformer \citep{melnychuk2022causaltransformer}. The treatment, outcome, and
covariate streams are initialized as
\begin{equation}
    u^{a}_{i,t}=W_a\phi_4(a_{i,t-1}),\qquad
    u^{y}_{i,t}=W_y\tilde y_{i,t},\qquad
    u^{x}_{i,t}=W_x x_{i,t}.
\end{equation}
Transformer blocks apply causal attention within and across streams. The final
representation is passed to balanced treatment and outcome heads. The loss is
\begin{equation}
    \mathcal{L}_{\mathrm{CT}}
    =
    \frac{
    \sum_{i,t}m_{i,t}
    \left(\hat y_{i,t+1}-\tilde y_{i,t+1}\right)^2
    }{
    \sum_{i,t}m_{i,t}
    }
    +
    \lambda_a
    \frac{
    \sum_{i,t}m_{i,t}\,
    \mathrm{CE}(\hat a_{i,t},a_{i,t})
    }{
    \sum_{i,t}m_{i,t}
    }.
\end{equation}
During counterfactual rollout, future covariates are hidden and predicted
outcomes are fed back autoregressively.

\subsection{G-Transformer}
\label{app:baseline_gtransformer}

G-Transformer is a transformer-based neural $g$-computation baseline inspired by
\citet{xiong2024gtransformer}. It uses treatment, outcome, and covariate streams
with transformer attention, but predicts outcomes through a factual
$g$-computation head rather than an adversarial treatment-balancing head. After
the transformer stack, the representation is mapped to
\begin{equation}
    h^r_{i,t}=\operatorname{ELU}(W_r h_{i,t}+c_r),
\end{equation}
and the one-step head predicts $\tilde y_{i,t+1}$ from
$[h^r_{i,t},\phi_4(a_{i,t})]$. The loss is the masked factual MSE,
\begin{equation}
    \mathcal{L}_{\mathrm{GT}}
    =
    \frac{
    \sum_{i,t}m_{i,t}
    \left(\hat y_{i,t+1}-\tilde y_{i,t+1}\right)^2
    }{
    \sum_{i,t}m_{i,t}
    }.
\end{equation}
At test time, the one-step head is applied autoregressively under the planned
future treatment sequence.

\begin{table}[t]
  \centering
  \small
  \caption{Summary of baseline mechanisms. The baselines cover
  inverse-probability weighting, neural $g$-computation, adversarial balancing,
  and transformer-based longitudinal sequence modeling.}
  \label{tab:baseline_summary}
  \begin{tabular}{llll}
    \toprule
    \textbf{Method} & \textbf{Sequence model} & \textbf{Adjustment mechanism} & \textbf{Outcome objective} \\
    \midrule
    MSM & Linear / ridge regressors & IPTW via logistic propensities & Weighted direct-horizon regression \\
    RMSN & LSTM encoder--decoder & IPTW via RNN propensities & Weighted MSE \\
    G-Net & LSTM & Neural $g$-computation & Outcome/covariate MSE \\
    CRN & LSTM encoder--decoder & Gradient reversal & Outcome MSE $+$ treatment CE \\
    CT & Multi-input transformer & Gradient reversal & Outcome MSE $+$ treatment CE \\
    GT & Multi-input transformer & Neural $g$-computation & Factual outcome MSE \\
    \bottomrule
  \end{tabular}
\end{table}

\section{Evaluation protocol details}
\label{app:evaluation_protocol_details}

\paragraph{Rolling-origin filtering and indexing.}
The raw benchmark generators use trajectory length $T=60$ and projection horizon
$H=5$. They generate candidate rolling-origin rows using the generator-level
minimum-origin setting, currently $\texttt{min\_t\_obs}=10$ in the benchmark
configuration. The shared evaluation layer uses the same minimum observed
history length and applies common validity filters: only rows with
\begin{equation}
    t_{\mathrm{obs}}\ge 10,\qquad
    t_{\mathrm{obs}}\le 64,\qquad
    t_{\mathrm{target}}\le 65
\end{equation}
are scored. Thus, $10$ is the minimum observed history length for both generated
candidate origins and reported evaluation rows in the current configuration.

One-step rows use
\begin{equation}
    t_{\mathrm{obs}}=\texttt{sequence\_lengths}-1,
    \qquad
    t_{\mathrm{target}}=\texttt{sequence\_lengths}.
\end{equation}
Horizon-5 rows use the stored rolling-origin index with
\begin{equation}
    t_{\mathrm{obs}}=\texttt{patient\_current\_t}+1,
    \qquad
    t_{\mathrm{target}}=t_{\mathrm{obs}}+5.
\end{equation}
These conventions match the inclusive observation-time convention used in
Section~\ref{sec:problem_formulation}: the query history is observed through
$t_{\mathrm{obs}}$, the first planned treatment is $a_{t_{\mathrm{obs}}}$, and
the target is $Y_{t_{\mathrm{target}}}$.

\paragraph{Support-only normalization and clipping.}
Outcome normalization statistics are computed from the support trajectories of
each benchmark file. Query targets are never used to estimate normalization
statistics. The reported normalized target is
\begin{equation}
    Y^{\mathrm{eval}}_{i,t}
    =
    \clip\!\left(
        \frac{Y_{i,t}-\mu_Y^{\mathrm{eval}}}{\sigma_Y^{\mathrm{eval}}},
        -10,10
    \right),
\end{equation}
where $(\mu_Y^{\mathrm{eval}},\sigma_Y^{\mathrm{eval}})$ are computed from the
full support set. Reported predictions are expressed in the same evaluation
normalization and clipped to $[-20,20]$:
\begin{equation}
    \widehat Y^{\mathrm{eval}}_{i,t}
    =
    \clip\!\left(
        \frac{\widehat Y^{\mathrm{raw}}_{i,t}-\mu_Y^{\mathrm{eval}}}
             {\sigma_Y^{\mathrm{eval}}},
        -20,20
    \right).
\end{equation}
For \CausalLongPFN{}, the model may internally predict in the PFN-context
normalization. Before scoring, predictions are converted back through the
context outcome scale and then into the shared evaluation normalization.

The reported normalized RMSE is
\begin{equation}
    \mathrm{RMSE}_{\mathrm{norm}}
    =
    \sqrt{
    \frac{1}{|\mathcal I_{\mathrm{eval}}|}
    \sum_{(i,t)\in\mathcal I_{\mathrm{eval}}}
    \left(
        \widehat Y^{\mathrm{eval}}_{i,t}
        -
        Y^{\mathrm{eval}}_{i,t}
    \right)^2
    },
\end{equation}
where $\mathcal I_{\mathrm{eval}}$ denotes the set of scored query rows for the
corresponding dataset, task, and method.


\section{Reproducibility, statistical uncertainty, and compute}
\label{app:reproducibility_compute}

This appendix summarizes the artifacts, evaluation protocol, statistical
aggregation, and compute assumptions needed to reproduce the experiments in
Section~\ref{sec:experiments}. The model architecture is specified in
Appendix~\ref{app:arch}, the synthetic TSCM prior in Appendix~\ref{app:prior},
the loss and training procedure in Appendices~\ref{app:loss}--\ref{app:training},
the rollout protocol in Appendix~\ref{app:rollout}, the evaluation datasets in
Appendix~\ref{app:evaluation_datasets}, the baseline models in
Appendix~\ref{app:baseline_models}, and the shared scoring protocol in
Appendix~\ref{app:evaluation_protocol_details}.

\paragraph{Released artifacts.}
The implementation, synthetic episode generator, training configuration, rollout
code, benchmark-construction scripts, baseline runners, and evaluation scripts
are available at
\url{https://github.com/Amirhossein-Zare/causal-long-pfn}. The pretrained
inference-only \CausalLongPFN{} weights used for frozen-model evaluation are
released on the Hugging Face Model Hub at
\url{https://huggingface.co/Amirhossein-Zare/causal-long-pfn}. The reported
\CausalLongPFN{} checkpoint uses the architecture and optimization
hyperparameters in Tables~\ref{tab:hyperparams} and~\ref{tab:optim_hparams} and
is pretrained entirely on synthetic episodes generated online from the TSCM prior
in Appendix~\ref{app:prior}. No target-domain trajectories are used during
\CausalLongPFN{} pretraining.

\paragraph{Reproducing benchmark evaluations.}
Cancer, warfarin, and HIV are simulated or semi-mechanistic domains with
branchable counterfactual labels, as described in
Appendix~\ref{app:evaluation_datasets}. These datasets can be regenerated from
the released code using the simulator specifications, task-grid configuration,
and random seeds. For these domains, query labels are obtained by replaying the
same patient-specific dynamics under the evaluated treatment sequence. MIMIC-III
is a credentialed-access de-identified clinical database and cannot be
redistributed with this paper; reproducing MIMIC-III results requires access
through the official data-use process, the preprocessing protocol described in
Appendix~\ref{app:data_mimic}, the released evaluation code, and the released
pretrained weights. MIMIC-III evaluation is factual rolling-origin prediction
only.

\paragraph{Baseline reproducibility.}
All baselines are trained using only target-domain support trajectories.
Hyperparameters are selected by support-set validation using the search spaces in
Table~\ref{tab:baseline_hparam_search_spaces}. The evaluation runner uses
grouped tuning: an initial random search is performed for the first dataset in
each $(\text{domain}, n_{\mathrm{sup}})$ group, and the best cached candidate is
reused and re-evaluated for later datasets in that group. The selected
configuration is then refit on the full support set before query evaluation.
Query outcomes are never used for hyperparameter selection. Thus, the baselines
receive domain-specific training and validation-based model selection, whereas
\CausalLongPFN{} is evaluated as a single frozen pretrained model without
test-time parameter updates, validation-based model selection, or target-domain
hyperparameter tuning.

\paragraph{Statistical uncertainty.}
The aggregation unit is one benchmark task: a fixed method, domain, support
size, task-index level, random repetition, and prediction horizon. For each
unit, normalized RMSE is computed after aggregating all scored query rows within
that task. Domain-level results report the mean across benchmark tasks. When
standard errors are reported, they are computed as
\begin{equation}
    \operatorname{SE}(\hat m)
    =
    \frac{\operatorname{sd}(m_1,\ldots,m_J)}{\sqrt{J}},
\end{equation}
where $m_j$ is the normalized RMSE for benchmark task $j$ and $J$ is the number
of tasks in the aggregation. Domain-balanced summaries are computed by first
averaging within each domain and then averaging the resulting domain means
equally across domains. For MIMIC-III, these summaries describe factual
prediction variability and not uncertainty in individual counterfactual effects.

\paragraph{Compute resources.}
The reported \CausalLongPFN{} pretraining run uses batch size $16$ with gradient
accumulation over $16$ micro-batches, giving an effective batch size of $256$
synthetic episodes, and is configured for $10{,}000$ optimizer steps. The
implementation supports CUDA training and multi-GPU data parallelism when
multiple GPUs are available. Baseline methods require additional compute because they are trained and
selected separately for target-domain benchmark tasks.

\section{Data assets, licenses, and ethics}
\label{app:data_assets_ethics}

The proposed \CausalLongPFN{} model, synthetic TSCM prior, and generated
synthetic training episodes are new research assets introduced by this work. The
paper documents their intended use, causal assumptions, limitations,
architecture, training procedure, rollout protocol, and evaluation protocol in
Sections~\ref{sec:methods}--\ref{sec:limitations_broader_impact} and
Appendices~\ref{app:prior}--\ref{app:reproducibility_compute}. Code for the model, synthetic data generation, benchmark construction,
evaluation, and pretrained-weight loading is available at
\url{https://github.com/Amirhossein-Zare/causal-long-pfn}. The pretrained
inference-only \CausalLongPFN{} weights are available from the Hugging Face
Model Hub at
\url{https://huggingface.co/Amirhossein-Zare/causal-long-pfn}.

The cancer, HIV, and warfarin benchmarks, the MIMIC-III dataset, and the
baseline methods are based on previously published benchmarks, simulators,
datasets, or modeling frameworks cited in
Appendices~\ref{app:evaluation_datasets} and~\ref{app:baseline_models}. This
paper credits the original sources used for benchmark construction and baseline
comparison. Reused simulator code, preprocessing code, baseline
implementations, and released model weights should be used in accordance with
their respective licenses and terms of use.

MIMIC-III is a de-identified credentialed-access clinical database and is not
redistributed with this paper. Reproducing MIMIC-III experiments requires
obtaining access through the official data-use process and applying the
preprocessing protocol described in Appendix~\ref{app:data_mimic}. Results on
MIMIC-III are factual rolling-origin prediction results and should not be
interpreted as validation of individual counterfactual treatment effects under
unobserved ICU interventions.

This work does not involve new human-subject recruitment, prospective
interventions, crowdsourcing, or direct interaction with patients. The clinical
component uses an existing de-identified database, and all reported MIMIC-III
results are aggregate benchmark metrics. \CausalLongPFN{} should be viewed as a
research tool for causal sequence modeling and hypothesis generation, not as a
standalone clinical decision system.

\end{document}